\documentclass[sigconf, nonacm]{acmart}
\renewcommand\footnotetextcopyrightpermission[1]{} 
\AtBeginDocument{%
 }

\makeatletter
\gdef\@copyrightpermission{
  \begin{minipage}{0.3\columnwidth}
   \href{https://creativecommons.org/licenses/by/4.0/}{\includegraphics[width=0.90\textwidth]{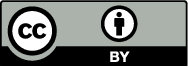}}
  \end{minipage}\hfill
  \begin{minipage}{0.7\columnwidth}
   \href{https://creativecommons.org/licenses/by/4.0/}{This work is licensed under a Creative Commons Attribution International 4.0 License.}
  \end{minipage}
  \vspace{5pt}
}
\makeatother

\usepackage{xcolor}
\usepackage{multirow}
\usepackage{booktabs}
\usepackage{hyperref}
\usepackage{seqsplit}
\usepackage{tabularray}
\usepackage{enumitem}
\usepackage{tabularx}

\begin{document}

\title{Reward Shaping for Happier Autonomous Cyber Security Agents}

\author{Elizabeth Bates}
\email{ebates@turing.ac.uk}
\affiliation{%
  \institution{The Alan Turing Institute}
  \country{United Kingdom}
}

\author{Vasilios Mavroudis}
\email{vmavroudis@turing.ac.uk}
\affiliation{%
  \institution{The Alan Turing Institute}
  \country{United Kingdom}
}

\author{Chris Hicks}
\email{c.hicks@turing.ac.uk}

\affiliation{%
  \institution{The Alan Turing Insitiute}
  \country{United Kingdom}}



\begin{abstract}
As machine learning models become more capable, they have exhibited increased potential in solving complex tasks. One of the most promising directions uses deep reinforcement learning to train autonomous agents in computer network defense tasks. This work studies the impact of the reward signal that is provided to the agents when training for this task. Due to the nature of cybersecurity tasks, the reward signal is typically 1) in the form of penalties (e.g., when a compromise occurs), and 
2) distributed sparsely across each defense episode. Such reward characteristics
are atypical of classic reinforcement learning tasks where the agent is regularly rewarded for progress (cf. to getting occasionally penalized for failures). We investigate reward shaping techniques that could bridge this gap so as to enable agents to train more sample-efficiently and potentially converge to a better performance. We first show that deep reinforcement learning algorithms are sensitive to the magnitude of the penalties and their relative size. Then, we combine penalties with positive external rewards and study their effect compared to penalty-only training. Finally, we evaluate intrinsic curiosity as an internal positive reward mechanism and discuss why it might not be as advantageous for high-level network monitoring tasks. 
\end{abstract}







\settopmatter{printfolios=true}
\maketitle

\section{Introduction}
Advanced Persistent Threats pose a significant challenge for defenders as they 
involve various attack tactics and vectors over prolonged periods, thus impeding event correlation, detection, and mitigation. Defenders need to monitor the network
for abnormal behaviour and take immediate remediation actions when a system gets compromised or traces of an adversary are found. Such actions usually involve a 
large array of tools for monitoring, scanning and reverting systems in a benign
state. Similarly, the adversary needs to combine a variety of tools to perform
reconnaissance, gain a foothold, escalate their access and move laterally towards their goal (e.g., exfiltration, impact).
In this arms race, the adversary is at an advantage. They can gather information about the target system or network in advance, prepare their steps and rapidly move through the network or remain present stealthily in compromised systems for a long time. In contrast, the defender needs to be able to fend off all attacks quickly and patch the vulnerabilities that enabled the intrusion in the first place. At present most of the incidence response relies on human operators that handle events raised by monitoring software.

Deep reinforcement learning (DRL) is actively researched and has been shown to
excel in interactive tasks that cannot easily be solved using analytical solutions. For example, human and even super-human levels of performance have been reported
in a range of tasks including board, video and strategy games~\cite{mnih2016asynchronous, Mnih2015, mnih2013playing, schulman_proximal_2017},
as well as autonomous driving~\cite{sallab_deep_2017} and robotics~\cite{kober_reinforcement_2013}. In such RL environments, rewards are typically positive and provide a consistent signal to the agent of how well they are doing during training (e.g., point gathered in a game). 

In the context of cybersecurity, DRL has been successfully applied to autonomously defend computer networks~\cite{cage1_win,foley,candles, cage3aicd} which, as discussed above, is a highly interactive task where the defending agent needs to intervene promptly and efficiently. However, cybersecurity environments have certain unusual characteristics that make training more challenging. Similar to human operators, the agent is given only penalties (cf. rewards) when compromises occur or when an adversary impacts a critical system. Consequently, the reward signal is sparsely distributed and focuses on unwanted events (e.g., breaches) rather than proactive defense behaviors (e.g., scanning systems regularly).

This work studies the effect of these reward signal characteristics and investigates various \textit{reward shaping} techniques that could improve the final performance of the agent or/and its sample efficiency (i.e., training speed). Overall, reward shaping for autonomous network defense tasks is an under-explored area; our main contributions are:
\vspace{1ex}
\begin{itemize}[leftmargin=*]
    \item We investigate whether the relative magnitude of the penalties has an effect on sample efficiency or the performance of the agent.  
    
    \item We introduce positive rewards along the penalties of the environment and study their effect in comparison to a baseline (penalty-driven) agent, as well as the impact of their relative magnitude.
    
    \item We study curiosity, a sophisticated internal reward technique that addresses reward sparsity by motivating exploration intrinsically, and compare its performance to a non-curious RL algorithm.
\end{itemize}

Our trained agents and augmented environments will be available online under an open source license.

\section{Background}

\subsection{Reinforcement Learning}
Reinforcement Learning (RL) is a type of machine learning which aims to learn the optimal behaviour (known as the optimal policy) in a given environment through experience and subsequent rewards or penalties for particular actions taken, given their consequence. Sutton and Barto (2018)~\cite{sutton} describe this learning system as ``hedonistic'' in its approach, as the focus is to maximise a special signal from the environment. Advances in the RL field in the last decade have demonstrated the ability and potential of RL agents, beyond the traditional RL algorithms like Monte-Carlo or tabular Q-Learning. These earlier approaches, whilst able to converge to optimal behaviours, do so in computationally and time intensive ways, and are not scalable to the complexity of many relevant environments. The introduction of deep Learning into the RL space has allowed agents to learn increasingly complex policies in environments with extensive, continuous state spaces and actions~\cite{DBLP}. Deep RL has yielded successful algorithms such as Trust Region Policy Optimisation (TRPO)~\cite{schulman2015} and Proximal Policy Optimisation (PPO)~\cite{schulman_proximal_2017} and use of Deep Q-Networks (DQN)~\cite{DQN}.

\subsection{Proximal Policy Optimisation (PPO)}
Proximal Policy optimisation (PPO) is a policy gradient based method which has achieved much success across RL literature~\cite{openai2019dota, PPO, yu2022surprising}, and is considered state of the art alongside algorithms like Soft Actor-Critic (SAC)~\cite{haarnoja2018soft} and the use of DQNs~\cite{DQN}. Policy gradient based methods formulate an objective function such that its gradient is an estimator of the policy gradient. The objective function (Equation~\ref{eqn:obj_fun}) is defined as the expected rewards over a trajectory $\tau$, which is dependent upon the policy $\pi$ which is, in turn, dependent upon parameters $\theta$, also known as the network weights. This can be reformulated (Equation~\ref{eqn:dif_obj_fun}) and differentiated to be written to use the advantage function $(A_{\pi_\theta})$, which determines whether a certain action is better to take than another, given the agent is in a particular state.
\vspace{-1ex}

\begin{equation}
\label{eqn:obj_fun}
\begin{aligned}
J(\theta)=\mathbb{E}_{\tau \sim \pi_\theta} R(\tau)=\sum_\tau P(\tau ; \theta) R(\tau)
\end{aligned}
\end{equation}

\begin{equation}
\label{eqn:dif_obj_fun}
\begin{aligned}
\nabla_\theta J(\theta)=\mathbb{E}_{\pi_0}\left[\nabla_\theta \log \pi_\theta(s, a) A_{\pi_\theta}(s)\right] \\ \text{where } A_{\pi_\theta}(s)=Q_{\pi_\theta(s, a)}-V_{\pi_\theta}(s)
\end{aligned}
\end{equation}

PPO is an extension on the earlier work of Trust Region Policy Optimisation (TRPO)~\cite{schulman2015}, aiming to reduce the complexity of TRPO whilst retaining the reliable performance and data efficiency. TRPO and PPO both go on to maximise the objective function, but in a constrained way such the policy updated are not “destructively large”~\cite{schulman_proximal_2017}. TRPO uses hard constraints, whilst PPO uses a clipping method (Equation \ref{eqn:clip}) where the ratio of the old to new policy must not exceed 1 + $\epsilon$ or 1 – $\epsilon$, depending on the value of the advantage term. The Actor-Critic strategy ~\cite{schulman_proximal_2017} (Equations \ref{eqn:PPO}, \ref{eqn:adv_t}) can be introduced such that the actor models the policy and the critic models the state-value function in two Deep Neural Networks, and iteratively helps each network improve as training progresses.  

\begin{equation}
\label{eqn:clip}
\begin{aligned}
L^{C L I P}(\theta)=\hat{\mathbb{E}}_t\left[\min \left(r_t(\theta) \hat{A}_t, \operatorname{clip}\left(r_t(\theta), 1-\epsilon, 1+\epsilon\right) \hat{A}_t\right)\right]
\end{aligned}
\end{equation}

\begin{equation}
\label{eqn:PPO}
\begin{aligned}
L_t^{C L I P+V F+S}(\theta)=\hat{\mathbb{E}}_t\left[L_t^{C L I P}(\theta)-c_1 L_t^{V F}(\theta)+c_2 S\left[\pi_\theta\right]\left(s_t\right)\right],
\end{aligned}
\end{equation}

\begin{equation}
\label{eqn:adv_t}
\begin{aligned}
\hat{A}_t=-V\left(s_t\right)+r_t+\gamma r_{t+1}+\cdots+\gamma^{T-t+1} r_{T-1}+\gamma^{T-t} V\left(s_T\right)
\end{aligned}
\end{equation}

Where $L_t^{V F}$ is critic loss, $c_1$ is the critic coefficient, $c_2$ is the entropy coefficient, $S$ represents an entropy bonus and $\gamma$ is the discount factor.

\subsection{Reward Shaping}
When applying reinforcement learning techniques to solve a problem that has a sparse reward signal, it can often take the agent a long time to learn the optimal behaviour. This can be seen in tasks that are episodic in nature e.g., a reward is only given to the agent on reaching the final goal state. Reward shaping is the notion of adding small rewards along the agent's trajectory to encourage faster learning and convergence to the optimal policy~\cite{devidze2022exploration}. Deciding how to shape these intermediate rewards so that they support learning the best behaviour is critical and non-straighforward. If incorrectly applied, the agent could learn sub-optimal policies, and become effectively ‘distracted’ from the real objective of the task. There are a variety of reward shaping techniques used throughout the literature, such as potential-based~\cite{Wiewiora_2003}, count-based~\cite{gupta2022unpacking, bellemare2016unifying, tang2017exploration}, curiosity-based~\cite{pathak2017curiosity, houthooft2016vime, burda2018large} and distance-based~\cite{ng1999policy, trott2019keeping}. The effects of reward shaping can help improve the efficiency of RL agents during training and improve sample efficiency~\cite{gupta2022unpacking}. Sample efficiency in this context refers to "the number of time steps in which the algorithm does not select near-optimal actions"~\cite{sutton2018reinforcement}. Altering the external rewards of an environment is a form of extrinsic reward shaping. However, there is very little discussion (if any) on augmenting reward signal in specialized cyber defence environments. The environment used in this paper has a sparse and penalty-driven reward signal, and is thus suitable for exploring the use of intrinsic and extrinsic reward shaping strategies and their effects.

\subsection{Intrinsic Curiosity}
Intrinsic Curiosity is a mechanism for encouraging an RL agent to explore novel states, or states that it is less certain about. An Intrinsic Curiosity Module (ICM) was first introduced by Pathak et al. (2017) and involves adding a curiosity-driven intrinsic reward ($r_t^{(i)}$) to the agent’s reward signal alongside extrinsic ($r_t^{(e)}$) environmental rewards~\cite{pathak2017curiosity}. The sum of these two rewards ($r_t=\eta_e r_t^{(e)}+\eta_i r_t^{(i)}$) are to be maximised as the policy improves, and the weighting of either reward can be altered using $\eta_e$ and $\eta_i$, where $\eta_i$=1-$\eta_e$ and $\eta_i$, $\eta_e$ are between 0 and 1~\cite{mazzaglia2022curiositydriven}. The ICM consists of a two neural networks, the inverse model and the forward model. First, state $s$ and $s_{t+1}$ are fed into a feature embedding layer. The inverse model takes in feature representations of state $s$ and $s_{t+1}$, $\phi(s_{t})$ and $\phi(s_{t+1})$ and outputs the predicted action given a state $s_t$ and subsequent state $s_{t+1}$. The feature encoded state is provided as input for the forward model, as well as the action $a_t$ to predict the feature representation of the next state $\hat{\phi}\left(s_{t+1}\right)$. The difference between this prediction and the actual value of $\phi(s_{t+1})$ is the internal reward signal~\cite{ pathak2017curiosity} added to the RL algorithm, in the case of this paper, PPO.

\section{Threat Model}\label{sec:threat}
We assume an adversary who aims to breach critical infrastructure systems in a 
enterprise or production computer network. For example, such a scenario could involve
the computer network of a manufacturing facility with both employee computers
and critical systems controlling the production line. We assume a sophisticated 
persistent adversary that moves \textit{laterally} across the 
systems of the network by progressively compromising hosts, and can potentially 
have prior information about the topology of the network.

This modus operandi is compatible with advanced persistent threats (APTs) where adversaries with a foothold in the network, discover, exploit, access, and escalate their privileges
in neighbouring systems. Such adversaries typically gain the initial foothold through a non-critical host of the network by employing an indirect strategy (e.g., phishing emails, a spear-phishing text, voice deepfakes, or a watering hole attack)~\cite{mirsky2021creation,li2016study}. 
The scope of the defences we consider in this work begins after the adversary gains a foothold. Thereafter the adversary employs various tools to perform reconnaissance, gain unauthorized access to systems, escalate their privileges, bypass security controls and launch attacks against the liveness of critical servers.

The adversary is not able to bypass network traffic routing limitations and 
cannot access systems that are not directly linked to the system they currently reside in.
For example, if the network is compartmentalized into subnets $A$, $B$ and $C$ and $A$ and $C$ not directly connected, the adversary is not able to attacks $C$'s systems from systems in $A$.
Moreover, the adversary can gain a foothold only through systems that are connected to the Internet and are either vulnerable or used in such a way that an indirect attack might be applicable (e.g., receiving emails). In contrast, the adversary cannot gain a foothold through systems that are offline or non-vulnerable. 
\vspace{-2ex}


\section{CYBorg Environment}\label{sec:cyborg}
For our experiments, we use the CybORG environment that simulates (and emulates) an enterprise computer network~\cite{cage_cyborg_2022,cyborg_acd_2021}. CybORG is one 
of the most commonly used environments for training cyber defense agents~\cite{foley, Bridging, nyberg2023training, yang2022behaviourdiverse} and provides a simulator paired with an emulator. This is to address the reality gap, a generalization problem that occurs when training RL agents in a simulated environment. CybORG's emulator runs on Amazon Web Services (AWS) and was used to validate that all the actions, observations and state transitions are consistent between the simulator and the emulator~\cite{cyborg_acd_2021}. 

A CybORG \textit{scenario} defines the network topology, the subnets and firewalls of the network along with the systems included in each subnet, their type (e.g., user host, enterprise server) and the services each of them exposes. Each episode is played between two actors: a defensive ``blue'' agent and an attacker ``red agent''. The environment makes available a number of actions to each agent and they take turns in interacting with network. Table~\ref{tbl:actions} summarizes the individual actions. Each action can be applied on a host of the network. The ``Decoy'' action sets up a honeypot for the adversary in the selected host and has seven variants (DecoyApache, DecoyFemitter, DecoyHarakaSMPT, DecoySmss, DecoySSHD, DecoySvchost, DecoyTomcat) depending on the service it mimics. To increase realism, each action has some likelihood to fail even when applied in a valid case (e.g., restore action applied on a compromised system). The red agents can perform an exploit action, and must specify which service by which to exploit the host. These decoy services set up by the blue agent can reveal red agent behaviour as any attempt to exploit a host through a decoy-service immediately fails. 

\begin{table}
\centering
\caption{The available actions that can be taken by the Blue and Red agents in a CybORG scenario.}
\begin{tblr}{
  row{1} = {c},
  row{2} = {c},
  cell{1}{1} = {c=2}{},
  hline{1,8} = {-}{0.08em},
  hline{2-3} = {-}{},
}
\textbf{ Actions}                                                                  &                                                       \\
{\textbf{Blue Defensive~}\\\textbf{Agent~}}                                        & {\textbf{Red Adversarial }\\\textbf{Agent}}           \\
1. Monitor a host                                                                  & {1. Scan a subnet for\\hosts}                         \\
{2.~Analyse processes~on a\\given host}                                            & {2. Discover Network Services\\of~a host}             \\
{3. Remove Red access, \\given the red agent has not \\escalated their privileges} & 3. Exploit a service~on~a port                        \\
{4. Restore a host to its initial \\configuration}                                 & {4. Escalate privilege on a \\host}                   \\
5. Set up decoy services                                                           & {5. Disrupt the services on~the\\~operational server} 
\end{tblr}
\label{tbl:actions}
\vspace{-1ex}
\end{table}

The observation space available to the defender is probabilistic in nature and is a vector consisting of 52 bits, with 4 bits corresponding to each host on the network. The first 2 bits indicate if the host has been scanned or exploited by the adversary, and the remaining two bits represent what level of access the adversary has on that host, either none, user or administrator~\cite{cage_challenge_2,foley}. 

To measure the success of the defenders, the CybORG environment uses a scoring function. The scores are all negative, hence sometimes referred to as penalties. The defender receives these penalties every time the red agent gains administrator access to a system, with the magnitude varying depending on how critical that host is to the enterprise network. The possible negative scores range from -0.1 to -1. The blue agent also receives a penalty of -10 if the red agent successfully impacts the ``Operational Server'' (green system in the operational subnet in Figure~\ref{fig:network}). Overall, the penalties reflect the importance of the system in the enterprise network. Due to the disruption to benign user operations, the defender receives a penalty of -1 when they perform the ``restore'' action on any host. The relationship between scores and rewards is discussed further in Section~\ref{subsec:score_augmentation}.

To make training RL agents more straightforward, the environment provides a wrapper that realises an OpenAI Gym interface (\href{https://www.gymlibrary.dev/}{www.gymlibrary.dev}, \href{https://gymnasium.farama.org/}{gymnasium.farama.org}) enabling agents to act as attackers, defenders or both~\cite{cage_challenge_2_announcement, aiGym}. This interface is compatible with all major RL frameworks.

\section{The CAGE Challenge}\label{subsec:cage}
This CybORG environment is released and maintained by the TTCP CAGE Working Group and was used in all three of the ``Cyber Autonomy Gym for Experimentation'' (CAGE) challenges~\cite{cage_challenge_1, cage_challenge_2, cage_challenge_3_announcement}. Each CAGE challenge introduced a (gradually more complex) cyber defense scenario implemented on CybORG and lasted for a period of three months each. During this period teams train and submit defense agents competing for the best score. The goal of the competition is to incentivise further research in autonomous defense agents. The CAGE environments and scenarios are have been used in several past work on autonomous decision making in a computer network environment~\cite{wolk2022cage, foley, nyberg2023training, yang2022behaviourdiverse}.

In this work, we use the second CAGE scenario as it is the most recent challenge with a fixed network topology. In contrast, the third CAGE challenge uses a mesh network where transmission links are constantly changing. This version of the problem is more difficult and requires a multi-agent RL solution. So far no efficient RL agents have been published (for CAGE 3) to be used as a baseline for our experiments and the top-performing solutions for the challenge were based on heuristics. 

\begin{figure}[h]
  \centering
  \includegraphics[width=0.9\columnwidth]{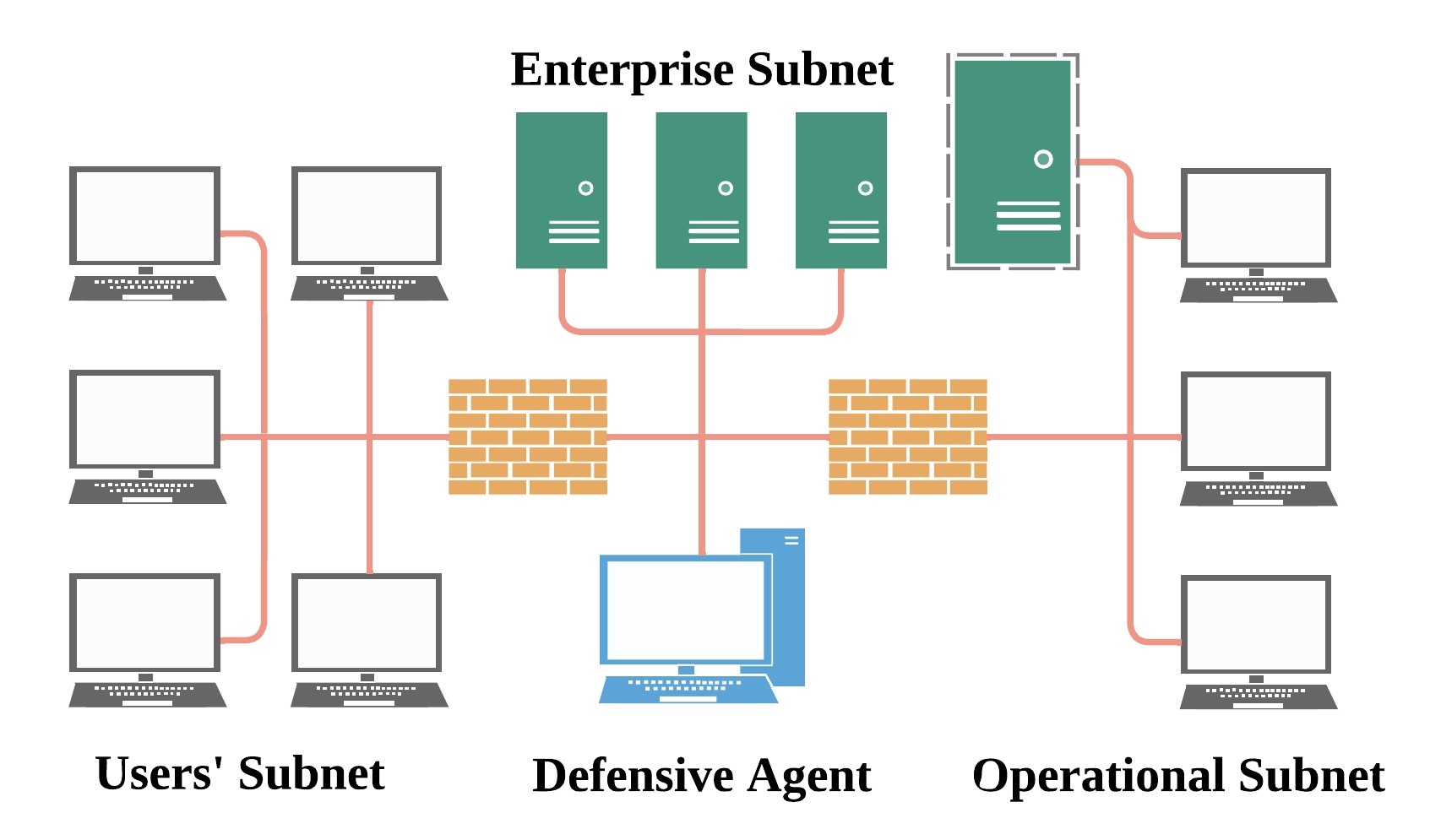}
  \caption{A visualisation of the CybORG Network topology~\cite{cage_challenge_2_announcement}, the network used in the CAGE challenge 1 and 2, featuring 3 subnets separated by firewalls. The green system in the operational subnet is the \textit{Operational Server} which is a critical system.}
  \vspace{-2ex}
  \label{fig:network}
\end{figure}

In the CAGE 2 scenario, the CybORG network is compartmentalized in three subnets, separated by firewalls (Figure~\ref{fig:network}): Subnet 1 consists of 5 user hosts, subnet 2 of three enterprise servers and subnet 3 of an operational server and 3 operational hosts. The firewalls prevent direct movement from subnet 1 to subnet 3, and thus an adversary can only get to subnet 3 via subnet 2. This topology is identical to the one used in the first of the iteration of the CAGE challenge. 

Each episode begins with the adversary (red agent) gaining a foothold on one of the user hosts in subnet 1. Subsequently, they can choose specific actions to move laterally through the network towards their goal which is the Operational Server (green system in operational subnet in Figure~\ref{fig:network}). This server maintains a service critical to the manufacturing facility that uses the enterprise network. Once the adversary has reached subnet 3, they seek to disrupt the operation of the Operational Server for as long as possible~\cite{cage_challenge_2} (Denial of Service attack). This setup and the goals of the adversary is compatible with the threat model introduced in Section~\ref{sec:threat}. 
\vspace{-3ex}

The challenge introduces also two scripted adversaries (with the same end-goal but different strategies) that each defender has to protect the network against: the ``meander'' and the ``Bline'' adversary.  \textit{Meander} has no prior knowledge of the network's topology, and thus thoroughly explores each subnet before moving to the next. In contrast, \textit{Bline} has prior information of the network's layout and follows an almost optimal (but randomized) trajectory towards the operational server. The defender (blue agent) has to take remediation (e.g., Remove action) and preventative (e.g., setting up decoy services) actions to contain the adversary. Note that full removal of the adversary from the foothold user host is not possible.

\subsection{CybORG Versions}\label{subsec:versions}
The CAGE Challenge 2 code base has been updated since the submission deadline of the competition on the 18th of July 2022. The updates included two separate bug fixes in August 2022 fixing: 1) an exploit failing on User Hosts 3 and 4, and 2) a bug in the initial observation of the agent that indicated an adversary having scanned on all hosts. The top-performing (state-of-the-art) solution~\cite{-cyborg-cage-2} was able to achieve their best scores using the version of the CAGE 2 challenge available at the time of submission (SHA: f12ff493cd3b8327cab93f645d71330e7b282377). However, for our experiments we use the newest version (commit on the 19th of August 2022) with the two bug fixes (SHA: 9421c8e985627810c5cac2abf0bcb62dfa6749fc).
Section~\ref{subsec:baseline} discusses further the impact of these patches to the pre-existing solutions and the steps we have taken to establish a reliable baseline.

\section{Considerations \& Methodology}\label{sec:methodology} 
Before we move on with our reward shaping experiments, we establish a baseline using a state-of-the-art solution from the literature. We use the winning agent of the CAGE 2 Challenge that combines the PPO Algorithm with greedy decoys~\cite{-cyborg-cage-2}. As we discuss in Sections~\ref{subsec:versions} and~\ref{subsec:baseline}, we retrain this agent to use the latest version of the environment. Thereafter, we refer to this retrained agent as \textit{baseline}.


\subsection{Score Terminology \& Augmentation}\label{subsec:score_augmentation}
In the context of this paper and the CAGE 2 Challenge environment, rewards refer to the individual numerical value given to the agent from the environment after taking an action, moving from one state to the next. The rewards available in the CAGE 2 environment were 0.0, -0.1, -1.0 and -10.0, depending on the action taken and importance of the servers or hosts compromised. Since all the rewards in the CAGE 2 Challenge environment are negative, we also refer to them as ``penalties''.
These baseline extrinsic rewards are altered in Experiment sets 1 and 2, and are therefore referred to as the ``augmented rewards''. Moreover, ``score'' refers to the cumulative rewards the defending agent acquires after a given amount of steps defending the network. In the CAGE challenge 2, this score was evaluated on episodes with 30, 50 and 100 steps.  

For our experiments, we introduced minor adjustments in the scoring function of CybORG. In particular, we exposed a method that alter the structure of the extrinsic rewards whilst also retaining the outputs of the original scoring function. In other words, at each step we output both the original rewards and the augmented reward from the reward shaping experiment. Tracking both of these values means that the agent can use the altered rewards to learn, whilst also being able to track what the original rewards would have been for the same defense strategy. For example, this allows us to plot the original rewards for the behaviour learnt from the altered rewards, so that the learning curves and the best average model scores can be easily compared to the baseline PPO model in the unmodified CAGE 2 challenge environment.

\subsection{State of the Art Defender}\label{subsec:baseline}
The baseline used in our experiments combines an Actor-Critic PPO agent and greedy decoys to achieve the best model scores (Table~\ref{tab:baselines_table}). Prior works had begun using a Dueling Deep Q-Network (DDQN), but replaced this with a PPO agent due to concerns of stability during training. However, the plain PPO implementation struggled to target decoys for specific hosts, in that potentially unhelpful decoy actions could be taken for that specific host (e.g., twice placing the same decoy, selecting a less effective decoy). This implementation included having a single decoy action selected greedily from all the possible decoys. Thus, an adjustment was made to change how decoy actions were chosen. The final solution included a greedy decoy placement strategy which greedily used only the most effective decoy from the nine available decoys for each host~\cite{-cyborg-cage-2}. The hyperparameters used for the baseline Actor-Critic PPO algorithm can be seen in Table~\ref{tbl:hyperparas} and these are used for all the experiments throughout this paper.



As discussed in Section~\ref{subsec:versions} the CAGE challenge solutions were 
trained on a flaw version of the environment. However, the performance of the baseline
agent was not reproducible in the patched version. To confirm that the discrepancy was due to the bug fixes in the code, we cloned a the competition version of the CAGE 2 Challenge repository and trained the baseline agent there. As seen in Table~\ref{tab:baselines_table}, 
the reported baseline results where then reproduced confirming that changes in the environment
affected the agent's performance. In our experiments, we use the most recent version of the CAGE 2 Challenge code base with the bugs fixed, and hence the new scores in the second row of Table~\ref{tab:baselines_table} are used as the baseline (SOTA) to compare reward shaping model scores against.

\begin{table*}
\caption{Scores of the top-performing defense agent in the competition and the patched versions of CAGE 2 Challenge for episodes of 30, 50 and 100 steps. There are notable differences in the Bline Adversary scores, but much less of a difference for the Meander Red Agent versions. The standard deviations were not included on the competition results page \cite{cage_challenge_2_announcement} and are denoted as N/A.}
\label{tab:baselines_table}
    \begin{tabular}[t]{ccccccccccccc}
        \toprule
        \multirow {4}{*}[-6ex]{\textbf{Experiment}} & \multicolumn{12}{c}{\textbf{Average scores and standard deviation}} \\
        \cmidrule{2-13} \\
        {} & \multicolumn{6}{c}{\textbf{Bline Red Agent}} & \multicolumn{6}{c}{\textbf{Meander Red Agent}} \\
        \cmidrule{2-7} \cmidrule{8-13}\\
        {} & \multicolumn{2}{c}{\textbf{30}} & \multicolumn{2}{c}{\textbf{50}} & \multicolumn{2}{c}{\textbf{100}} & \multicolumn{2}{c}{\textbf{30}} & \multicolumn{2}{c}{\textbf{50}} & \multicolumn{2}{c}{\textbf{100}}\\
        \cmidrule{2-3} \cmidrule{4-5} \cmidrule{6-7} \cmidrule{8-9} \cmidrule{10-11} \cmidrule{12-13}\\
        {} & \textbf{Score} & \textbf{$\sigma$} & \textbf{Score} & \textbf{$\sigma$} & \textbf{Score} & \textbf{$\sigma$} & \textbf{Score} & \textbf{$\sigma$} & \textbf{Score} & \textbf{$\sigma$} & \textbf{Score} & \textbf{$\sigma$} \\
        \midrule
        \textbf{Competition (f12ff49)} & -3.47 & N/A & -6.41 & N/A & -13.76 & N/A & -5.64 & N/A & -8.69 & N/A & -16.6 & N/A \\
        \textbf{Latest (9421c8e} & -4.232 & 2.247 & -7.596 & 3.190 & -15.993 & 5.491 & -5.624 & 1.345 & -8.894 & 2.224 & -16.996 & 4.285 \\
        \bottomrule
    \end{tabular}

\end{table*}

\subsection{Hardware, Algorithms \& Hyperparameters}
As discussed, we use the top-performing agent from CAGE 2 (i.e., baseline) as a starting point for our rewards shaping experiments. For training (actor-critic PPO) we use the Adam optimiser to update the actor and critic networks’ weights. PPO is a policy gradient method that is typically sample efficient and often matches or even outperforms other state of the art (SOTA) methods~\cite{schulman_proximal_2017, mnih2016asynchronous, Surjit_2020}. The specific hyperparameters used for PPO in all our experiments are listed in Table~\ref{tbl:hyperparas}. We used 5 Azure Standard D48s v3 (48 vcpus, 192 GiB memory) virtual machines to train our agents in parallel.

\begin{table}[h!]
\centering
\caption{Hyperparameters for the Actor-Critic PPO algorithm used in all our experiments.}
\begin{tblr}{
  hline{1,11} = {-}{0.08em},
  hline{2} = {-}{},
}
\textbf{Hyperparameters}                           & \textbf{Values}         \\
Learning rate ($\alpha$) & 0.002          \\
Epochs (K)                                & 6              \\
Minibatch Size (in timesteps)             & 20,000         \\
Discount ($\gamma$)                          & 0.99           \\
GAE parameters ($\lambda$)                   & 1.0            \\
Betas                                     & {[}0.0, 0.990] \\
Clipping Coefficent ($\epsilon$)             & 0.2            \\
$C_1$                                        & 0.5            \\
$C_2$                                        & 0.01           
\end{tblr}
\label{tbl:hyperparas}
\end{table}
\vspace{-2ex}

\section{Experimental Evaluation}\label{sec:experiments}
Three sets of experiments were conducted, covering both intrinsic and extrinsic reward shaping ideas. The first two cover extrinsic reward shaping ideas by exploring scaling up the magnitude of the rewards and adding a mix of positive and negative rewards to the environment. The third focuses on the results of applying an Intrinsic Curiosity Module (ICM) to explore how it affects the PPO learning agent in the CAGE 2 context. From these experiments, we aim to: 1) gain a deeper understanding of the effect of extrinsic rewards in a cyber environment, 2) make comparisons with the impact of intrinsic rewards, and 3) investigate whether reward shaping alterations can lead to improved sample complexity or a faster convergence to optimal behaviour. 

Due to the inherent variance in training of RL models, quantitative results are more reliable when averaged over several training sessions~\cite{stable-baselines}. In each experiment, the same model is trained from scratch 15 times for 75,000 episodes each. We then evaluate the scores and standard deviations of the trained models for 1000 episodes of 30, 50 and 100 steps. Although this is a large number of iterations, it matches the iterations used in the evaluation script of the CAGE 2 Challenge. There was no signficant improvement in models after 50,000 episodes of training, thus all the figures are plotted up to the 50,000 episode-mark.

Each experiment was conducted twice, against the Red Bline agent and then the Red Meander agent.
The third sleep agent was also evaluated and the scores were consistently 0, as expected, and thus we do not include this agent's results in any tables or figures. 
In all experiments a larger reward indicates better performance. Moreover, to examine the statistical significance between our experimental results we use the P-value~\cite{andrade2019p} between the performance means, and report accordingly on each experiment. 
Each experiment was run in parallel (Section~\ref{sec:methodology}), taking approximately 24 hours each. Our trained agents and augmented environments will be publicly available as open source.
\vspace{-1ex}

\subsection{Exp. 1: Extrinsic - Magnitude Change}
In this experiment, we study how the magnitude of the extrinsic penalties applied to a blue agent affects how sample-efficiently the agent learns to protect the network. The baseline penalties are the initial values set in the CAGE environment (see Section~\ref{subsec:score_augmentation}). We alter these and generate a new tuple of rewards with 1) the manually augmented penalties, and 2) the baseline penalties that the agent would receive for the same behaviour. Keeping both of these values means that the agent can learn from the augmented rewards, but the performance of the agents can be fairly compared on the basis of the same scoring function.

We perform three scaled-reward experiments: one with normalised rewards, one with rewards scaled up by one order of magnitude and one scaled up disproportionately to the importance of the system (any red action/presence on the operational server) rewards.  In the context of this experiment, the normalised set of augmented rewards means that the reward values are between 0 and -1, as opposed to that of the baseline CAGE environment which are between 0 and -10. See Table~\ref{tbl:mag_rew} for the augmented rewards in comparison to the CAGE baseline.

The main incentive for exploring adjusting the rewards in this manner are that preliminary experiments in the OpenAI Gym's Mountain Car environment demonstrated changes in the learning curve when similar experiments were trialled. The Mountain Car environment is relevant as it too is a sparse reward environment with mostly negative rewards, much like the CAGE 2 Environment. 

From our review of the literature, there are few works that perform a systematic investigation of the effects of reward magnitude.
Reward normalisation is a strategy used in other RL experiments, such as in the initial Deep Q-Learning (DQN) paper~\cite{DQN} which trained an RL agent to successfully play a multitude of Atari games. The rationale in~\cite{DQN} was to try to maintain the same hyperparameters of the agent when training across all Atari games (some games had much larger rewards than others). Another benefit is that clipping the rewards this way limited the scale of the error derivatives, which could potentially lead to more efficient training. This concept of large rewards leading to large error derivatives and potentially slowing the agent's training was also the reason that the largest negative reward for both experiment 2 (Scaled-up by one order of magnitude) and experiment 3 (disproportionately scaled-up) was capped at -100.

\begin{table}[h]
\centering
\caption{Reward intervals for the set of extrinsic reward shaping experiments on penalty magnitude change. Baseline rewards are the rewards used in the original CAGE 2 environment while the rest of the columns include augmented rewards.}
\begin{tabular*}{\columnwidth}{cccc}
\toprule
\multicolumn{4}{c}{\bfseries Reward Intervals} \\
\hline
\textbf{Baseline} & \textbf{Normalised} & \vtop{\hbox{\strut \textbf{Scaled-up by}}\hbox{\strut \textbf{1 order of }}\hbox{\strut \textbf{magnitude}}} & \vtop{\hbox{\strut \textbf{Scaled-up}}\hbox{\strut \textbf{disproportionately}}}\\
\hline
0.0     & 0.0  & 0.0  & 0.0\\
-0.1    & -0.01  & -1.0  & -0.1\\
-1.0    &  -0.1 & -10.0 & -10.0\\
-10.0    &  -1.0 & -100.0  & -100.0\\ 
\hline

\end{tabular*}

\label{tbl:mag_rew}
\vspace{-5ex}
\end{table}

\begin{figure}[h]
  \centering
  \includegraphics[width=\linewidth]{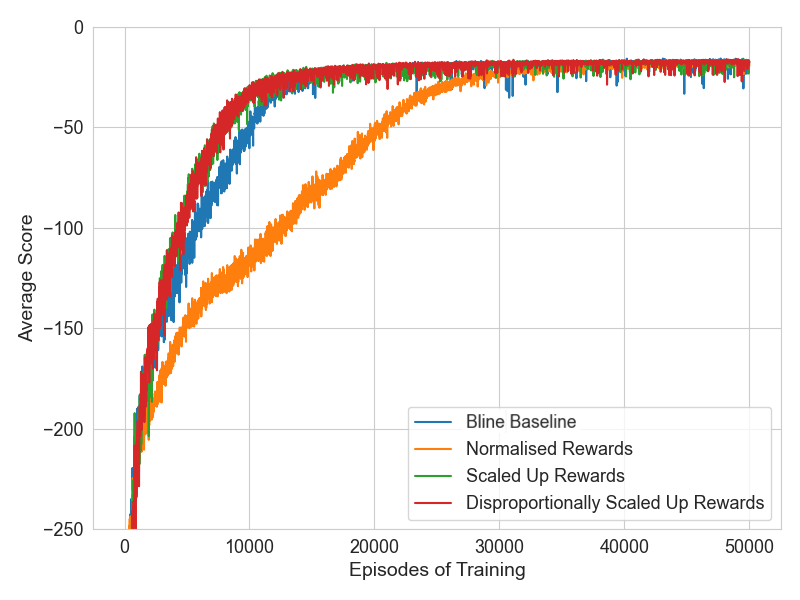}
  \caption{The average training curves for the Experiment set 1, where the attacker is the Bline Red agent. We conduct three magnitude-altering experiments using normalised, scaled-up by one order of magnitude and disproportionately scaled-up rewards. Each curve is the average of 15 models
  trained for 50,000 episodes. The two sets of scaled up rewards seem to achieve better scores than the baseline and the normalised rewards throughout the start of the training, then they all plateau at similar scores. Score in this context refers to the cumulative sum of rewards for 100 timesteps.}
  \label{fig:exp1_bline_fig}
  \vspace{-1ex}
\end{figure}

Figure~\ref{fig:exp1_bline_fig} and Table~\ref{tab:exp1_table} show that the two experiments with scaled up the rewards performed the best for the first 10,000 episodes of training. Both then converged to average scores of -16.132 and -15.680 for 100 episodes for the scaled-up by one order of magnitude rewards and the disproportionately scaled-up rewards experiments respectively. The disproportionately scaled-up rewards experiment achieved a marginally better average score (-15.993) than the baseline PPO agent. As seen in Figure~\ref{fig:exp1_bline_fig}, both experiments that augmented the rewards by scaling them up were able to achieve the same performance (the performance difference in both cases was not statistically significant, they had P-values of >0.95 for both experiments in comparison to the baseline score) with the baseline agent but in a more sample efficient manner.

In comparison, the learning curve of the normalised rewards experiment took significantly longer to achieve an average score (-15.946) that was not significantly different statistically than that of the  baseline -15.993 (for P=0.98). The slower convergence is likely due to the very small (and sparse) rewards that provide only an attenuated signal to the agent that is not significant enough to promote learning efficiently.

\begin{figure}[h]
  \centering
  \includegraphics[width=\linewidth]{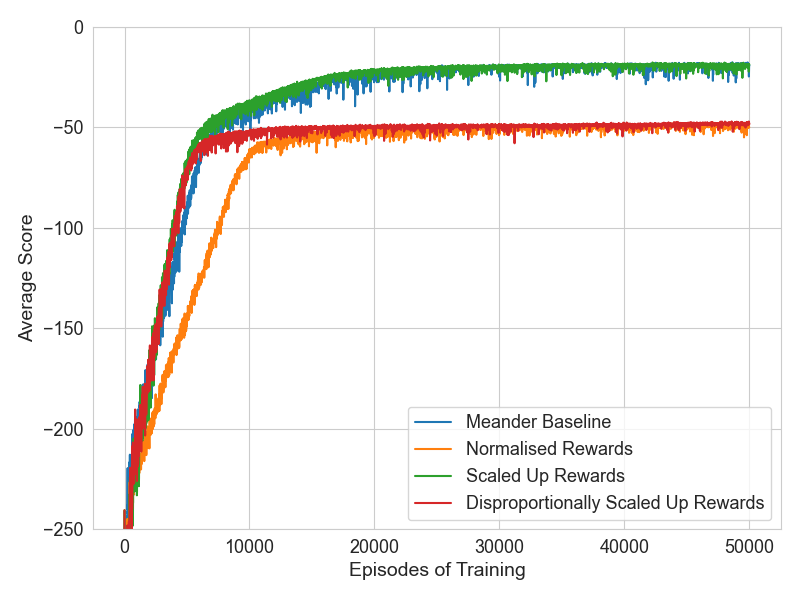}
  \caption{
  The average training curves for the Experiment set 1, where the attacker is the Meander Red agent. We examine three magnitude-altering strategies using normalised, scaled-up by one order of magnitude and disproportionately scaled-up rewards. Each curve is the average of 15 models trained with each set of altered rewards (for 50,000 episodes). In this case, only the scaled-up by one order of magnitude experiment performed equally to the baseline, and the rest of the experiments plateaued at much lower scores of approximately -43. Score in this context refers to the cumulative sum of rewards for 100 timesteps.}
  \label{fig:exp1_meander_fig}
  \vspace{-1ex}
\end{figure}

However, it can be seen in Figure~\ref{fig:exp1_meander_fig} and the same table that these trends were not consistent against the Red Meander agent. Both the normalised and the disproportionately scaled-up rewards experiments converged at scores much worse than the baseline result, at -43.695 and -42.210 for a 100 time step episode respectively compared to the baseline's -16.996. In contrast, 

the scaled-up rewards experiment was able to achieve a much better score of -17.237. All the experiments, except the normalised one, performed similarly up until approximately 6,000 episodes of training, which is where the disproportionately scaled-up rewards experiment began to plateau, whilst the baseline and scaled-up to one order of magnitude rewards continued to improve, but at a slower rate. As with the Bline agent, the normalised rewards agents took the longest to converge to its best average score.

\newcolumntype{C}[1]{>{\centering\let\newline\\\arraybackslash\hspace{0pt}}m{#1}}

\begin{table*}[h]
\caption{Average scores and standard deviations for Experiment set 1 for 30, 50 and 100 steps of the trained blue agent defending against either the Bline red agent or the Meander red agent. Average baseline scores for models trained only with PPO and no augmented rewards are included for comparison. The PPO baseline remains the best average score for attack from the Meander Red Agent, however the differences in average scores were not statistically significant to the baseline scores for any experiments, except for the normalised and disproportionately scaled-up rewards experiments.}

\label{tab:exp1_table}
\centering
\begin{tabular}[t]{C{3cm}cccccccccccc}
        \toprule
        \multirow {4}{*}{\textbf{Experiment}} & \multicolumn{12}{c}{\textbf{Average scores and standard deviation}} \\
        \cmidrule{2-13} \\
        {} & \multicolumn{6}{c}{\textbf{Bline Red Agent}} & \multicolumn{6}{c}{\textbf{Meander Red Agent}} \\
        \cmidrule{2-7} \cmidrule{8-13}\\
        {} & \multicolumn{2}{c}{\textbf{30}} & \multicolumn{2}{c}{\textbf{50}} & \multicolumn{2}{c}{\textbf{100}} & \multicolumn{2}{c}{\textbf{30}} & \multicolumn{2}{c}{\textbf{50}} & \multicolumn{2}{c}{\textbf{100}}\\
        \cmidrule{2-3} \cmidrule{4-5} \cmidrule{6-7} \cmidrule{8-9} \cmidrule{10-11} \cmidrule{12-13}\\
        {} & \textbf{Score} & \textbf{$\sigma$} & \textbf{Score} & \textbf{$\sigma$} & \textbf{Score} & \textbf{$\sigma$} & \textbf{Score} & \textbf{$\sigma$} & \textbf{Score} & \textbf{$\sigma$} & \textbf{Score} & \textbf{$\sigma$} \\
        \midrule
        \textbf{Baseline} & -4.232 & 2.247 & -7.596 & 3.190 & -15.993 & 5.491 & -5.624 & 1.345 & -8.894 & 2.224 & -16.996 & 4.285 \\
        
        \textbf{Normalised rewards} & -4.253 & 2.177 & -7.591 & 3.088 & -15.946 & 5.271 & -7.077 & 1.458 & -16.995 & 2.642 & -43.695 & 4.760\\

        \textbf{Scaled-up rewards} & -4.247 & 2.255 & -7.652 & 3.254 & -16.132 & 5.690 & -5.630 & 1.359 & -8.893 & 2.231 & -17.237 & 8.202\\

        \textbf{Disproportionately scaled-up rewards} & -4.179
        & 2.135 & -7.473 & 3.009 & -15.680 & 4.875 & -6.814 & 1.450 & -16.267 & 2.681 & -42.210 & 6.544 \\
        \bottomrule
    \end{tabular}

\end{table*}

\subsection{Exp. 2 - Extrinsic - Positive Rewards}
We now explore introducing positive rewards in an otherwise entirely penalty-driven learning problem. Two sets of models were trained, one with a small positive reward of 0.1 added, and the second with a larger positive reward of 1.0. The positive reward was introduced by adding a reward for all the times the agent would usually receive a reward signal of 0.0, i.e. when the defensive blue agent is performing actions proactively but those reveal no adversarial presence. See Table~\ref{tbl:positive_rew} for the positive augmented rewards in comparison to the CAGE baseline. 
Similarly with Experiment 1, our motivation was our preliminary experiments with positive and negative rewards in the simpler mountain car environment as well as the reward shaping techniques introduced in~\cite{magnitude_paper}. More specifically, \cite{magnitude_paper} investigated limiting their rewards between 0 and 8, and between -8 and 8 such that both successes and failures were rewarded/penalised proportionally. This notion of including both positive and negative rewards was an interesting idea and although applied in a different environment, it seemed like a relevant avenue to explore, especially since these negative rewards are set in the CAGE 2 environment and have not been studied previously.

\begin{table}[h]
\caption{Reward Intervals for the set of extrinsic reward shaping experiments on adding positive rewards. Both the experiments augment the reward value that is usually 0.0 in the baseline CAGE 2 challenge environment, to a small positive reward of 0.1 or a larger reward of 1.0.}

\centering
\begin{tabular*}{\columnwidth}{ccc}
\toprule
\multicolumn{3}{c}{\bfseries Reward Intervals} \\
\hline
\textbf{Baseline}
& \textbf{Small positive reward}
& \textbf{Large positive reward}\\
\hline
0.0     & 0.1   & 1.0 \\
-0.1    & -0.1  & -0.1\\
-1.0    & -1.0  & -1.0\\
-10.0   & -10.0 & -10.0\\ 
\hline
\end{tabular*}
\label{tbl:positive_rew}
\vspace{-2ex}
\end{table}

The results of these two experiments in Table~\ref{tab:exp2_table} show that the addition of both small and large positive rewards does minorly improve upon baseline results for the Bline agent, though this improvement is not statistically significant. The addition of the small rewards does achieve a score for 30 timesteps of -4.072 which is better than the baseline's -4.232. Moreover, the experiments outperform the baseline at 50 and 100 time steps; the baseline at 50 was -7.596 and the agent's was -7.352 and for 100 the baseline was -15.993 against the agent's -15.849. All the standard deviations were also smaller than the baselines across all the Bline agent at 30, 50 and especially at 100, indicating more reliable performance when small positive rewards are added to this environment.

For scores against the red Meander agent, the addition of the small positive rewards produced scores identical to the baseline agent (statistically non-significant difference), while the addition of the larger positive rewards performed worse than the baseline, especially for the 100 timesteps evaluation.

By adding a positive reward for states where no red adversaries are encountered, an additional positive incentive is provided to keep the network healthy (in addition to the environment's penalties). Moreover, this positive addition makes the environment's rewards less sparse i.e., the agent is now receiving relevant rewards much more frequently. Finally, this shows that rewarding the absence and penalising the presence of the adversary \textit{did not} incentivise the agent to game the rewards by e.g., avoiding scanning potentially compromised systems.

\begin{figure}[h]
  \centering
  \includegraphics[width=\linewidth]{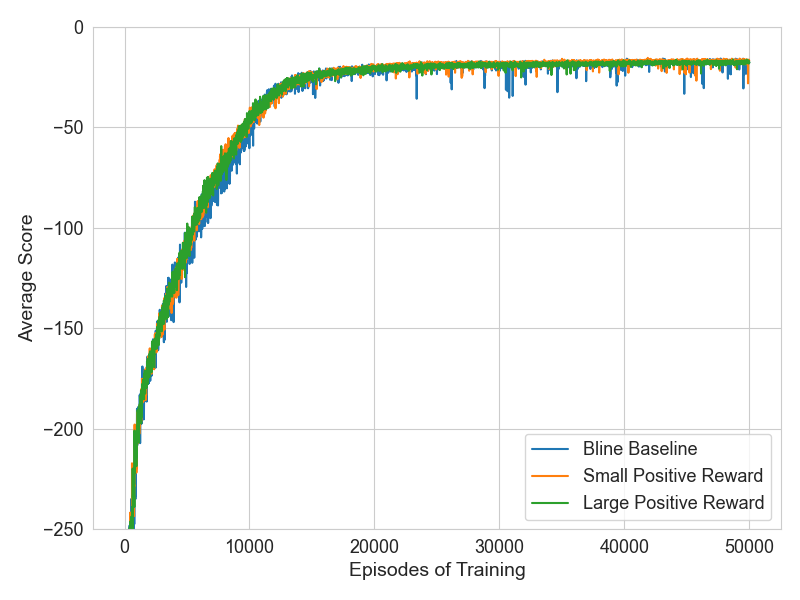}
  \caption{
  The average training curves for the Experiment set 2, where the attacker is the Bline Red agent. Each curve is the average of 15 models trained with each set of altered rewards (for 50,000 episodes). The learning curves for both the small and large addition of positive rewards experiments are very similar in rate of score improvement as the baseline curve, though both experiments converge to slightly better values (see Table~\ref{tab:exp2_table}). Score in this context refers to the cumulative sum of rewards for 100 timesteps.}
  \label{fig:exp2_bline_fig}
\end{figure}

\begin{figure}[h]
  \centering
  \includegraphics[width=\linewidth]{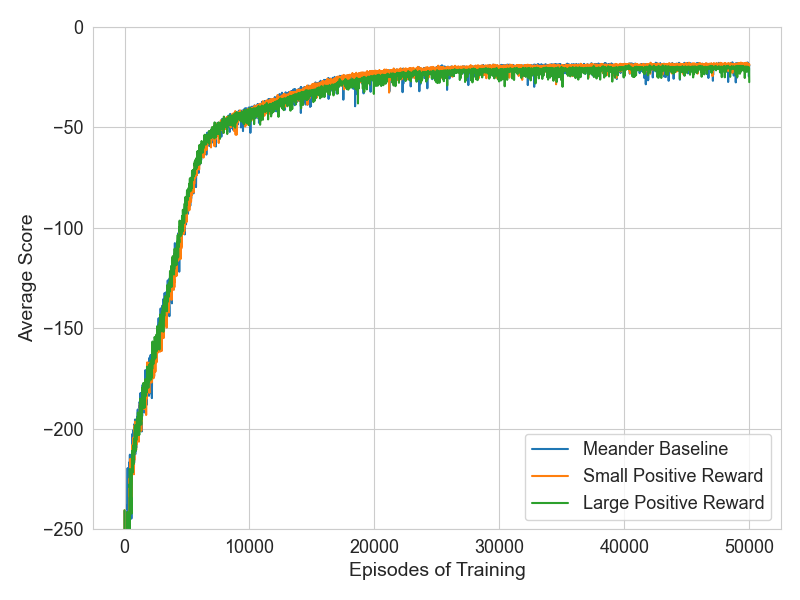}
  \caption{
  The average training curves for the Experiment set 2, where the attacker is the Meander Red agent. Each curve is the average of 15 models trained with each set of altered rewards (for 50,000 episodes). The two sets of scaled up rewards seem to achieve better scores than the baseline and the normalised rewards throughout the start of the training, then they all converge to similar scores. 
  Again, the learning curves for both the small and large addition of positive rewards experiments are very similar in rate of score improvement as the baseline curve, though they do converge to perform slightly worse than the baseline.
  }
  \label{fig:exp2_meander_fig}
\end{figure}

\begin{table*}[h]
\caption{Average scores and standard deviations for Experiment set 2 for 30, 50 and 100 steps of the trained blue agent defending against either the Bline red agent and the Meander red agent. Average baseline scores for models trained only with PPO and no augmented rewards are included for comparison. 
The scores that are statistically significantly better are in bold. 
}
\label{tab:exp2_table}
\centering
\begin{tabular}[t]{C{3cm}cccccccccccc}
        \toprule
        \multirow {4}{*}{\textbf{Experiment}} & \multicolumn{12}{c}{\textbf{Average scores and standard deviation}} \\
        \cmidrule{2-13} \\
        {} & \multicolumn{6}{c}{\textbf{Bline Red Agent}} & \multicolumn{6}{c}{\textbf{Meander Red Agent}} \\
        \cmidrule{2-7} \cmidrule{8-13}\\
        {} & \multicolumn{2}{c}{\textbf{30}} & \multicolumn{2}{c}{\textbf{50}} & \multicolumn{2}{c}{\textbf{100}} & \multicolumn{2}{c}{\textbf{30}} & \multicolumn{2}{c}{\textbf{50}} & \multicolumn{2}{c}{\textbf{100}}\\
        \cmidrule{2-3} \cmidrule{4-5} \cmidrule{6-7} \cmidrule{8-9} \cmidrule{10-11} \cmidrule{12-13}\\
        {} & \textbf{Score} & \textbf{$\sigma$} & \textbf{Score} & \textbf{$\sigma$} & \textbf{Score} & \textbf{$\sigma$} & \textbf{Score} & \textbf{$\sigma$} & \textbf{Score} & \textbf{$\sigma$} & \textbf{Score} & \textbf{$\sigma$} \\
        \midrule
        \textbf{Baseline} & -4.232 & 2.247 & -7.596 & 3.190 & -15.993 & 5.491 & -5.624 & 1.345 & -8.894 & 2.224 & -16.996 & 4.285 \\
        
        \textbf{Small positive rewards} & -4.072 & 2.033 & -7.352 & 2.967 & -15.457 & 4.955 & -5.640 & 1.354 & -8.890 & 2.210 & -17.005 & 5.776\\

        \textbf{Large positive rewards} & -4.210 & 2.210 & -7.475 & 3.128 & -15.849 & 5.111 & -5.700 & 1.369 & -8.963 & 2.270 & -17.205 & 6.429\\

        \bottomrule
    \end{tabular}

\end{table*}

\subsection{Exp. 3 - Intrinsic - ICM}\label{subsec:exp3}
This experiment studies intrinsic reward shaping, thus all the extrinsic rewards are reset back to their initial values used in the baseline models, ranging from 0.0, -0.1, -1.0 and -10.0. An additional model was created using an identical architecture to the Actor-Critic PPO model used in all the previous experiments with the addition of an Intrinsic Curiosity Module (ICM) incorporated into it. We retained the hyperparameters as reported in Section~\ref{tbl:hyperparas} and set five additional ones specific ICM (Table~\ref{tbl:icm_hp}).
$\beta$ was set to the value recommended in the initial Intrinsic Curiosity paper~\cite{pathak2017curiosity} while $\eta$, $\alpha_i$, $\eta_e$ and $\eta_i$ were set to match the configuration of mainstream implementations of a PPO agent with ICM~\footnote{\label{fn1}\href{https://github.com/adik993/ppo-pytorch/tree/master}{https://github.com/adik993/ppo-pytorch}}~\footnote{\href{https://github.com/chagmgang/pytorch_ppo_rl/tree/master}{https://github.com/chagmgang/pytorch{\textunderscore}ppo{\textunderscore}rl}}~\cite{mazzaglia2022curiositydriven}.

\begin{figure}[h]
  \centering
  \includegraphics[width=\linewidth]{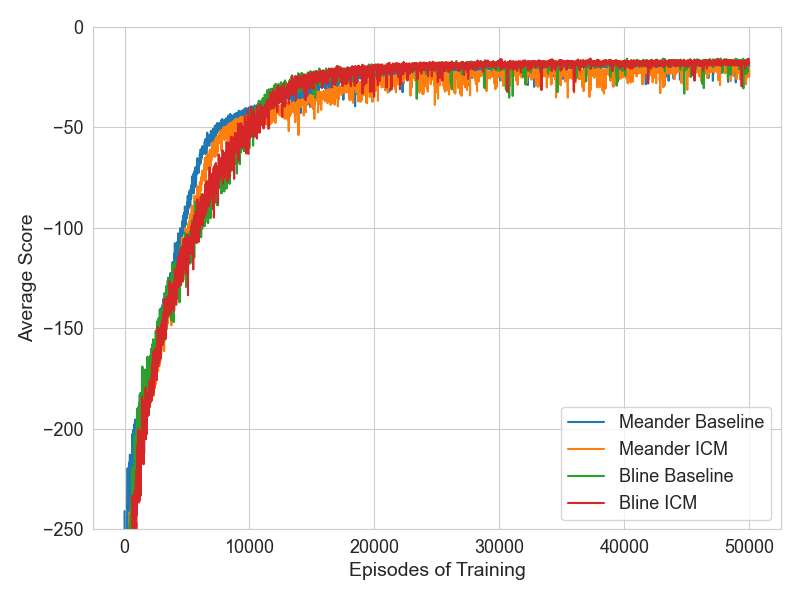}
  \caption{
  The average training curves for the Experiment 3, with both the Red Meander and Bline agent curves plotted. Each curve is the average of 15 models trained with each set of altered rewards (for 50,000 episodes). Score in this context refers to the cumulative sum of rewards for 100 timesteps. The inclusion of ICM has not significantly affected the learning speed and has lower average scores throughout training than the baseline PPO agent could achieve against either the red Bline or Meander agent.}
  \label{fig:exp3_fig}
  \vspace{-2ex}
\end{figure}

\vspace{-1ex}
As seen in Figure~\ref{fig:exp3_fig}, the inclusion of ICM does not improve the PPO agent's training speed or final performance, it in fact performs worse than the baseline agents. The standard deviation for each set of scores was larger for both the red Bline and Meander agent for almost all the sets of steps. This can be seen again in Figure~\ref{fig:exp3_fig} in the ICM learning curves that show much more variance throughout training. This larger variance could be explained by the greater emphasis on exploration that the ICM agents have, especially in the early stages of training. 
This larger variance indicates that baseline PPO agent can achieve scores close to the average more reliably. 

The lack of positive effect from introducing internal rewards through ICM is not entirely unexpected. It is likely that the baseline agent has been able to learn all defense strategies that were within its capabilities (given its observation space and the capacity of the policy neural network) and thus ICM could not help uncover new strategies. Moreover, curiosity has reportedly aided the Bline PPO agent in the CAGE Challenge 1 environment~\cite{cage1_win} and in the CAGE 2 environment prior to the submission deadline~\cite{foley}. The subsequent bug fixes mentioned in~\ref{subsec:versions} and the addition of the greedy decoys could have made curiosity redundant. In particular, minimising the need for more thorough exploration (in CAGE 2) and ruling out potentially unorthodox defense strategies that took advantage of the implementation bugs (in CAGE 1). 
Interestingly,~\cite{foley} also found that curiosity-based internal rewards did not benefit the defender's performance against the Meander agent. 

Finally, to ensure that the magnitude of the rewards ($\eta$) was not the impeding ICM from assisting, we ran a small scale experiment (single model, 30,000 steps) with  $\eta=1$. 
In comparison, an $\eta$ of $0.01$ is commonly used in RL frameworks\footnote{\href{https://github.com/adik993/ppo-pytorch/tree/master}{https://github.com/adik993/ppo-pytorch}} and when applying curiosity
in environments of similar reward sparsity with exclusively negative rewards e.g., Mountain Car~\footnote{\href{https://www.gymlibrary.dev/environments/classic_control/mountain_car/}{https://www.gymlibrary.dev/environments/classic{\textunderscore}control/mountain{\textunderscore}car}}~\cite{aiGym}. Even after this substantial amplification of the rewards, ICM did not improve neither the sample efficiency nor the final performance of the baseline model.

\begin{table}[h]
\centering
\caption{Additional hyperparameters table for the ICM used in Experiment 3.}
\begin{tblr}{
  hline{1, 7} = {-}{0.08em},
  hline{2} = {-}{},
}
\textbf{Hyperparameters}          & \textbf{Values} \\
ICM Learning rate ($\alpha_i$)    & 0.001  \\
ICM beta ($\beta_i$)              & 0.2    \\
Reward scale ($\eta$)             & 0.01   \\
External Reward Factor ($\eta_e$) & 0.9    \\
Internal Reward Factor ($\eta_i$) & 0.1    
\end{tblr}
\label{tbl:icm_hp}
\end{table}

\begin{table*}[]
\caption{Average scores and standard deviations for Experiment 3 ICM for 30, 50 and 100 steps of the trained blue agent defending against either the Bline red agent or the Meander red agent. Average baseline scores for models trained only with PPO and no augmented rewards are included for comparison.}
\label{tab:exp3_table}
\begin{tabular}[t]{ccccccccccccc}
        \toprule
        \multirow {4}{*}{\textbf{Experiment}} & \multicolumn{12}{c}{\textbf{Average scores and standard deviation}} \\
        \cmidrule{2-13} \\
        {} & \multicolumn{6}{c}{\textbf{Bline Red Agent}} & \multicolumn{6}{c}{\textbf{Meander Red Agent}} \\
        \cmidrule{2-7} \cmidrule{8-13}\\
        {} & \multicolumn{2}{c}{\textbf{30}} & \multicolumn{2}{c}{\textbf{50}} & \multicolumn{2}{c}{\textbf{100}} & \multicolumn{2}{c}{\textbf{30}} & \multicolumn{2}{c}{\textbf{50}} & \multicolumn{2}{c}{\textbf{100}}\\
        \cmidrule{2-3} \cmidrule{4-5} \cmidrule{6-7} \cmidrule{8-9} \cmidrule{10-11} \cmidrule{12-13}\\
        {} & \textbf{Score} & \textbf{$\sigma$} & \textbf{Score} & \textbf{$\sigma$} & \textbf{Score} & \textbf{$\sigma$} & \textbf{Score} & \textbf{$\sigma$} & \textbf{Score} & \textbf{$\sigma$} & \textbf{Score} & \textbf{$\sigma$} \\
        \midrule
        \textbf{Baseline} & -4.232 & 2.247 & 7.596 & 3.190 & -15.993 & 5.491 & -5.624 & 1.345 & -8.894 & 2.224 & -16.996 & 4.285 \\
        \textbf{ICM} & -4.483 & 2.576 & -8.297 & 4.368 & -18.127 & 9.990 & -6.030 & 1.503 & -9.742 & 2.843 & -18.967 & 7.807 \\
        \bottomrule
    \end{tabular}
    
\end{table*}

\section{Related Work}
The application of ML for autonomous cyber operations is still an emerging field. The capabilities and potential of autonomous agents trained using techniques such as RL enable defending networks autonomously at scale and speed, thus providing new efficient methods to preserve network security.  

To train such agents, various network simulation and emulation environments have been proposed in the literature e.g., CyberBattleSim~\cite{msft}, FARLAND~\cite{molina}, CANDLES ~\cite{candles}and Galaxy~\cite{kevin}. With the exception of Galaxy~\cite{kevin}, all these environments allow for parallelisation which significantly speeds training up. Candles and CyberBattleSim are strictly simulations of cyber security environments and use a finite state machine~\cite{cyborg_acd_2021}, while the others are emulations that use virtual machines (e.g., on the cloud). CyberBattleSim and FARLAND are the most recently published (2021) alongside CybORG. CyberBattleSim, much like CybORG, is based on the Open AI Gym environment and provides a high-level simulated abstraction of computer networks~\cite{msft}. FARLAND is a framework designed for training autonomous agents for network defense by gradually increasing complexity. 

Galaxy is similar framework, offering a modular environment to experiment and tailor to the requirements and constraints of different real-world systems. Unfortunately, Galaxy is limited in terms of scaling their network emulation; currently each host computer can support only one visualised network due to emulation being computationally expensive, restricting its ability to scale up. The developers of this framework are planning to containerise the infrastructure in future work to allow for higher-fidelity emulation and scalability~\cite{kevin}.

CybORG is one of the most commonly used environments in the literature. This is primarily due to the CAGE Challenges. Researchers, practitioners and participants have been publishing solutions and algorithm combinations demonstrating a variety of successful RL uses specifically to explore the creation of autonomous network defenders, both from a single-agent (CAGE 1,2~\cite{cage_challenge_1, cage_challenge_2}) and a multi-agent (CAGE 3) perspective. In particular, there is relevant work in the winning submission for the CAGE 1 Challenge~\cite{cage1_win} and 3rd place submission for the CAGE 2 Challenge. Both submissions included curiosity as an additional reward shaping strategy, which improved the scores of their CAGE 1 Bline PPO agent “by nearly double”~\cite{cage1_win} and significantly improved performance of their Bline agent in CAGE 2~\cite{foley} as well. Their CAGE 2 implementation showed that the solely PPO agent was able to achieve maximum best scores of approximately -10, and the two alternative agents trained were able to achieve much better scores between approximately -2 and -4~\cite{foley}. 

Similarly, CAGE scenarios (running on CybORG) have been used in various works within the field of autonomous cyber defense, such as that of Wolk et al. (2022)~\cite{wolk2022cage} which evaluates a set of different RL approaches, such as ensemble RL, action masking, hierarchical RL and custom training in cyberdefense scenarios. They selected CAGE due to its minimal reality gap when emulating an attacker on an enterprise network~\cite{wolk2022cage}. Furthermore, CAGE has been used in exploring automatic penetration testing~\cite{yang2022behaviourdiverse} where it was used as an environment to test the RL framework 'CLAP' alongside another SOTA penetration testing environment for networks, NASim~\cite{schwartz2019autonomous, nasim}.

Reward shaping for a reinforcement learning is a well-studied area, especially in environments with typically sparse rewards~\cite{ grzes2017reward, marom2018belief, ng1999policy}. Janssen and Grey’s (2012)~\cite{magnitude_paper} work explored different magnitudes of reward shaping in a cognition-based task called ‘Blocks World’. The magnitude of rewards (based on accuracy and time taken to complete the task) were altered to either be between 0 and 8, or -8 and 8. 
The results in their particular environment show that using either of these two reward bounds did not result in significant difference for their model in terms of accuracy~\cite{magnitude_paper}. Another example of deliberately altered reward values is~\cite{Singh} that simulated two agent environments and repeatedly varies the reward values within the range of -1 and 1. This work was not specifically focused on the effect of varying extrinsic rewards but instead discussed rewards in an evolutionary context~\cite{Singh}. Besides these, there is very limited research (especially in the cybersecurity context) on the impact of reward magnitude, combinations of negative and positive rewards and how they compare to baseline results and intrinsic reward strategies (such as ICM). 
\vspace{-2ex}

\section{Conclusion}
Cybersecurity environments differ significantly
from typical RL environments as the objective of the agent
is to preserve the initial (i.e., non-compromised) state of the system/network (i.e., any deviation from that results in a negative reward). This work took a first step in better understanding
how this characteristic affects learning and what techniques
can be used to train a performant agent more efficiently.
Our findings show that reward shaping can be effective in 
increasing sample efficiency and performance. However, depending
on the technique used and the magnitude of the rewards, the 
improvement can vary and may even result in performance decrease 
in some cases. In the future, such techniques could be combined with curriculum learning to quickly bootstrap learning (on simpler but relevant reward-augmented tasks) and then gradually fall back to non-augmented rewards on the original task.
\vspace{-1ex}

\begin{acks}
Research funded by the Defence Science and Technology Laboratory (Dstl) which is an executive agency of the UK Ministry of Defence providing world class expertise and delivering cutting-edge science and technology for the benefit of the nation and allies. The research supports the Autonomous Resilient Cyber Defence (ARCD) project within the Dstl Cyber Defence Enhancement programme.
\end{acks}


\bibliographystyle{ACM-Reference-Format}
\balance
\bibliography{references}


\begin{thebibliography}{56}


\ifx \showCODEN    \undefined \def \showCODEN     #1{\unskip}     \fi
\ifx \showDOI      \undefined \def \showDOI       #1{#1}\fi
\ifx \showISBNx    \undefined \def \showISBNx     #1{\unskip}     \fi
\ifx \showISBNxiii \undefined \def \showISBNxiii  #1{\unskip}     \fi
\ifx \showISSN     \undefined \def \showISSN      #1{\unskip}     \fi
\ifx \showLCCN     \undefined \def \showLCCN      #1{\unskip}     \fi
\ifx \shownote     \undefined \def \shownote      #1{#1}          \fi
\ifx \showarticletitle \undefined \def \showarticletitle #1{#1}   \fi
\ifx \showURL      \undefined \def \showURL       {\relax}        \fi
\providecommand\bibfield[2]{#2}
\providecommand\bibinfo[2]{#2}
\providecommand\natexlab[1]{#1}
\providecommand\showeprint[2][]{arXiv:#2}

\bibitem[cag(2021)]%
        {cage_challenge_1}
 \bibinfo{year}{2021}\natexlab{}.
\newblock \bibinfo{title}{Cyber Autonomy Gym for Experimentation Challenge 1}.
\newblock
  \bibinfo{howpublished}{\url{https://github.com/cage-challenge/cage-challenge-1}}.
\newblock
\newblock
\shownote{Created by Maxwell Standen, David Bowman, Son Hoang, Toby Richer,
  Martin Lucas, Richard Van Tassel}.


\bibitem[cyb(2021)]%
        {cyborg_acd_2021}
 \bibinfo{year}{2021}\natexlab{}.
\newblock \bibinfo{booktitle}{\emph{CybORG: A Gym for the Development of
  Autonomous Cyber Agents}}. \bibinfo{publisher}{arXiv}.
\newblock


\bibitem[cag(2022a)]%
        {cage_challenge_2}
 \bibinfo{year}{2022}\natexlab{a}.
\newblock \bibinfo{title}{Cyber Autonomy Gym for Experimentation Challenge 2}.
\newblock
  \bibinfo{howpublished}{\url{https://github.com/cage-challenge/cage-challenge-2}}.
\newblock
\newblock
\shownote{Created by Maxwell Standen, David Bowman, Son Hoang, Toby Richer,
  Martin Lucas, Richard Van Tassel, Phillip Vu, Mitchell Kiely}.


\bibitem[cag(2022b)]%
        {cage_cyborg_2022}
 \bibinfo{year}{2022}\natexlab{b}.
\newblock \bibinfo{title}{Cyber Operations Research Gym}.
\newblock
  \bibinfo{howpublished}{\url{https://github.com/cage-challenge/CybORG}}.
\newblock
\newblock
\shownote{Created by Maxwell Standen, David Bowman, Son Hoang, Toby Richer,
  Martin Lucas, Richard Van Tassel, Phillip Vu, Mitchell Kiely, KC C., Natalie
  Konschnik, Joshua Collyer}.


\bibitem[Andrade(2019)]%
        {andrade2019p}
\bibfield{author}{\bibinfo{person}{Chittaranjan Andrade}.}
  \bibinfo{year}{2019}\natexlab{}.
\newblock \showarticletitle{The P value and statistical significance:
  misunderstandings, explanations, challenges, and alternatives}.
\newblock \bibinfo{journal}{\emph{Indian journal of psychological medicine}}
  \bibinfo{volume}{41}, \bibinfo{number}{3} (\bibinfo{year}{2019}),
  \bibinfo{pages}{210--215}.
\newblock


\bibitem[Applebaum et~al\mbox{.}(2022)]%
        {Bridging}
\bibfield{author}{\bibinfo{person}{Andy Applebaum}, \bibinfo{person}{Camron
  Dennler}, \bibinfo{person}{Patrick Dwyer}, \bibinfo{person}{Marina
  Moskowitz}, \bibinfo{person}{Harold Nguyen}, \bibinfo{person}{Nicole
  Nichols}, \bibinfo{person}{Nicole Park}, \bibinfo{person}{Paul Rachwalski},
  \bibinfo{person}{Frank Rau}, \bibinfo{person}{Adrian Webster}, {and}
  \bibinfo{person}{Melody Wolk}.} \bibinfo{year}{2022}\natexlab{}.
\newblock \showarticletitle{Bridging Automated to Autonomous Cyber Defense:
  Foundational Analysis of Tabular Q-Learning}. In
  \bibinfo{booktitle}{\emph{Proceedings of the 15th ACM Workshop on Artificial
  Intelligence and Security}} (Los Angeles, CA, USA)
  \emph{(\bibinfo{series}{AISec'22})}. \bibinfo{publisher}{Association for
  Computing Machinery}, \bibinfo{address}{New York, NY, USA},
  \bibinfo{pages}{149–159}.
\newblock
\showISBNx{9781450398800}
\urldef\tempurl%
\url{https://doi.org/10.1145/3560830.3563732}
\showDOI{\tempurl}


\bibitem[Bellemare et~al\mbox{.}(2016)]%
        {bellemare2016unifying}
\bibfield{author}{\bibinfo{person}{Marc Bellemare}, \bibinfo{person}{Sriram
  Srinivasan}, \bibinfo{person}{Georg Ostrovski}, \bibinfo{person}{Tom Schaul},
  \bibinfo{person}{David Saxton}, {and} \bibinfo{person}{Remi Munos}.}
  \bibinfo{year}{2016}\natexlab{}.
\newblock \showarticletitle{Unifying count-based exploration and intrinsic
  motivation}.
\newblock \bibinfo{journal}{\emph{Advances in neural information processing
  systems}}  \bibinfo{volume}{29} (\bibinfo{year}{2016}).
\newblock


\bibitem[Brockman et~al\mbox{.}(2016)]%
        {aiGym}
\bibfield{author}{\bibinfo{person}{Greg Brockman}, \bibinfo{person}{Vicki
  Cheung}, \bibinfo{person}{Ludwig Pettersson}, \bibinfo{person}{Jonas
  Schneider}, \bibinfo{person}{John Schulman}, \bibinfo{person}{Jie Tang},
  {and} \bibinfo{person}{Wojciech Zaremba}.} \bibinfo{year}{2016}\natexlab{}.
\newblock \bibinfo{title}{OpenAI Gym}.
\newblock
\newblock
\showeprint{arXiv:1606.01540}


\bibitem[Burda et~al\mbox{.}(2018)]%
        {burda2018large}
\bibfield{author}{\bibinfo{person}{Yuri Burda}, \bibinfo{person}{Harri
  Edwards}, \bibinfo{person}{Deepak Pathak}, \bibinfo{person}{Amos Storkey},
  \bibinfo{person}{Trevor Darrell}, {and} \bibinfo{person}{Alexei~A Efros}.}
  \bibinfo{year}{2018}\natexlab{}.
\newblock \showarticletitle{Large-scale study of curiosity-driven learning}.
\newblock \bibinfo{journal}{\emph{arXiv preprint arXiv:1808.04355}}
  (\bibinfo{year}{2018}).
\newblock


\bibitem[CAGE(2022)]%
        {cage_challenge_2_announcement}
\bibfield{author}{\bibinfo{person}{CAGE}.} \bibinfo{year}{2022}\natexlab{}.
\newblock \bibinfo{title}{{TTCP CAGE Challenge 2}}.
\newblock
  \bibinfo{howpublished}{\url{https://github.com/cage-challenge/cage-challenge-2}}.
\newblock


\bibitem[Devidze et~al\mbox{.}(2022)]%
        {devidze2022exploration}
\bibfield{author}{\bibinfo{person}{Rati Devidze}, \bibinfo{person}{Parameswaran
  Kamalaruban}, {and} \bibinfo{person}{Adish Singla}.}
  \bibinfo{year}{2022}\natexlab{}.
\newblock \showarticletitle{Exploration-Guided Reward Shaping for Reinforcement
  Learning under Sparse Rewards}.
\newblock \bibinfo{journal}{\emph{Advances in Neural Information Processing
  Systems}}  \bibinfo{volume}{35} (\bibinfo{year}{2022}),
  \bibinfo{pages}{5829--5842}.
\newblock


\bibitem[Foley et~al\mbox{.}(2022)]%
        {cage1_win}
\bibfield{author}{\bibinfo{person}{Myles Foley}, \bibinfo{person}{Chris Hicks},
  \bibinfo{person}{Kate Highnam}, {and} \bibinfo{person}{Vasilios Mavroudis}.}
  \bibinfo{year}{2022}\natexlab{}.
\newblock \showarticletitle{Autonomous Network Defence Using Reinforcement
  Learning}. In \bibinfo{booktitle}{\emph{Proceedings of the 2022 ACM on Asia
  Conference on Computer and Communications Security}} (Nagasaki, Japan)
  \emph{(\bibinfo{series}{ASIA CCS '22})}. \bibinfo{publisher}{Association for
  Computing Machinery}, \bibinfo{address}{New York, NY, USA},
  \bibinfo{pages}{1252–1254}.
\newblock
\showISBNx{9781450391405}
\urldef\tempurl%
\url{https://doi.org/10.1145/3488932.3527286}
\showDOI{\tempurl}


\bibitem[Foley et~al\mbox{.}(2023)]%
        {foley}
\bibfield{author}{\bibinfo{person}{Myles Foley}, \bibinfo{person}{Mia Wang},
  \bibinfo{person}{Zoe M}, \bibinfo{person}{Chris Hicks}, {and}
  \bibinfo{person}{Vasilios Mavroudis}.} \bibinfo{year}{2023}\natexlab{}.
\newblock \bibinfo{title}{Inroads into Autonomous Network Defence using
  Explained Reinforcement Learning}.
\newblock
\newblock
\showeprint[arxiv]{2306.09318}~[cs.CR]


\bibitem[Group(2022)]%
        {cage_challenge_3_announcement}
\bibfield{author}{\bibinfo{person}{TTCP CAGE~Working Group}.}
  \bibinfo{year}{2022}\natexlab{}.
\newblock \bibinfo{title}{TTCP CAGE Challenge 3}.
\newblock
  \bibinfo{howpublished}{\url{https://github.com/cage-challenge/cage-challenge-3}}.
\newblock


\bibitem[Grzes(2017)]%
        {grzes2017reward}
\bibfield{author}{\bibinfo{person}{Marek Grzes}.}
  \bibinfo{year}{2017}\natexlab{}.
\newblock \showarticletitle{Reward shaping in episodic reinforcement learning}.
\newblock  (\bibinfo{year}{2017}).
\newblock


\bibitem[Gupta et~al\mbox{.}(2022)]%
        {gupta2022unpacking}
\bibfield{author}{\bibinfo{person}{Abhishek Gupta}, \bibinfo{person}{Aldo
  Pacchiano}, \bibinfo{person}{Yuexiang Zhai}, \bibinfo{person}{Sham Kakade},
  {and} \bibinfo{person}{Sergey Levine}.} \bibinfo{year}{2022}\natexlab{}.
\newblock \showarticletitle{Unpacking reward shaping: Understanding the
  benefits of reward engineering on sample complexity}.
\newblock \bibinfo{journal}{\emph{Advances in Neural Information Processing
  Systems}}  \bibinfo{volume}{35} (\bibinfo{year}{2022}),
  \bibinfo{pages}{15281--15295}.
\newblock


\bibitem[Haarnoja et~al\mbox{.}(2017)]%
        {DBLP}
\bibfield{author}{\bibinfo{person}{Tuomas Haarnoja}, \bibinfo{person}{Haoran
  Tang}, \bibinfo{person}{Pieter Abbeel}, {and} \bibinfo{person}{Sergey
  Levine}.} \bibinfo{year}{2017}\natexlab{}.
\newblock \showarticletitle{Reinforcement Learning with Deep Energy-Based
  Policies}.
\newblock \bibinfo{journal}{\emph{CoRR}}  \bibinfo{volume}{abs/1702.08165}
  (\bibinfo{year}{2017}).
\newblock
\showeprint[arXiv]{1702.08165}
\urldef\tempurl%
\url{http://arxiv.org/abs/1702.08165}
\showURL{%
\tempurl}


\bibitem[Haarnoja et~al\mbox{.}(2018)]%
        {haarnoja2018soft}
\bibfield{author}{\bibinfo{person}{Tuomas Haarnoja}, \bibinfo{person}{Aurick
  Zhou}, \bibinfo{person}{Pieter Abbeel}, {and} \bibinfo{person}{Sergey
  Levine}.} \bibinfo{year}{2018}\natexlab{}.
\newblock \bibinfo{title}{Soft Actor-Critic: Off-Policy Maximum Entropy Deep
  Reinforcement Learning with a Stochastic Actor}.
\newblock
\newblock
\showeprint[arxiv]{1801.01290}~[cs.LG]


\bibitem[Hannay(2022)]%
        {-cyborg-cage-2}
\bibfield{author}{\bibinfo{person}{John Hannay}.}
  \bibinfo{year}{2022}\natexlab{}.
\newblock \bibinfo{title}{{Cyborg Cage 2 Solution}}.
\newblock
  \bibinfo{howpublished}{\url{https://github.com/john-cardiff/-cyborg-cage-2}}.
\newblock


\bibitem[Hicks et~al\mbox{.}(2023)]%
        {cage3aicd}
\bibfield{author}{\bibinfo{person}{Chris Hicks}, \bibinfo{person}{Vasilios
  Mavroudis}, \bibinfo{person}{Myles Foley}, \bibinfo{person}{Thomas Davies},
  \bibinfo{person}{Kate Highnam}, {and} \bibinfo{person}{Tim Watson}.}
  \bibinfo{year}{2023}\natexlab{}.
\newblock \showarticletitle{{Canaries and Whistles: Resilient Drone
  Communication Networks with (or without) Deep Reinforcement Learning}}.
\newblock \bibinfo{journal}{\emph{{Proceedings of the 16th ACM Workshop on
  Artificial Intelligence and Security (AISec ’23), Copenhagen, Denmark}}}
  (\bibinfo{year}{2023}).
\newblock


\bibitem[Hill et~al\mbox{.}(2018)]%
        {stable-baselines}
\bibfield{author}{\bibinfo{person}{Ashley Hill}, \bibinfo{person}{Antonin
  Raffin}, \bibinfo{person}{Maximilian Ernestus}, \bibinfo{person}{Adam
  Gleave}, \bibinfo{person}{Anssi Kanervisto}, \bibinfo{person}{Rene Traore},
  \bibinfo{person}{Prafulla Dhariwal}, \bibinfo{person}{Christopher Hesse},
  \bibinfo{person}{Oleg Klimov}, \bibinfo{person}{Alex Nichol},
  \bibinfo{person}{Matthias Plappert}, \bibinfo{person}{Alec Radford},
  \bibinfo{person}{John Schulman}, \bibinfo{person}{Szymon Sidor}, {and}
  \bibinfo{person}{Yuhuai Wu}.} \bibinfo{year}{2018}\natexlab{}.
\newblock \bibinfo{title}{Stable Baselines}.
\newblock
  \bibinfo{howpublished}{\url{https://github.com/hill-a/stable-baselines}}.
\newblock


\bibitem[Houthooft et~al\mbox{.}(2016)]%
        {houthooft2016vime}
\bibfield{author}{\bibinfo{person}{Rein Houthooft}, \bibinfo{person}{Xi Chen},
  \bibinfo{person}{Yan Duan}, \bibinfo{person}{John Schulman},
  \bibinfo{person}{Filip De~Turck}, {and} \bibinfo{person}{Pieter Abbeel}.}
  \bibinfo{year}{2016}\natexlab{}.
\newblock \showarticletitle{Vime: Variational information maximizing
  exploration}.
\newblock \bibinfo{journal}{\emph{Advances in neural information processing
  systems}}  \bibinfo{volume}{29} (\bibinfo{year}{2016}).
\newblock


\bibitem[Janssen and Gray(2012)]%
        {magnitude_paper}
\bibfield{author}{\bibinfo{person}{Christian~P. Janssen} {and}
  \bibinfo{person}{Wayne~D. Gray}.} \bibinfo{year}{2012}\natexlab{}.
\newblock \showarticletitle{When, What, and How Much to Reward in Reinforcement
  Learning-Based Models of Cognition}.
\newblock \bibinfo{journal}{\emph{Cognitive Science}} \bibinfo{volume}{36},
  \bibinfo{number}{2} (\bibinfo{year}{2012}), \bibinfo{pages}{333--358}.
\newblock
\urldef\tempurl%
\url{https://doi.org/10.1111/j.1551-6709.2011.01222.x}
\showDOI{\tempurl}
\showeprint{https://onlinelibrary.wiley.com/doi/pdf/10.1111/j.1551-6709.2011.01222.x}


\bibitem[Kober et~al\mbox{.}(2013)]%
        {kober_reinforcement_2013}
\bibfield{author}{\bibinfo{person}{Jens Kober}, \bibinfo{person}{J.~Andrew
  Bagnell}, {and} \bibinfo{person}{Jan Peters}.}
  \bibinfo{year}{2013}\natexlab{}.
\newblock \showarticletitle{Reinforcement learning in robotics: {A} survey}.
\newblock \bibinfo{journal}{\emph{International Journal of Robotics Research}}
  \bibinfo{volume}{32} (\bibinfo{date}{Sept.} \bibinfo{year}{2013}),
  \bibinfo{pages}{1238--1274}.
\newblock
\showISSN{0278-3649}
\urldef\tempurl%
\url{https://doi.org/10.1177/0278364913495721}
\showDOI{\tempurl}


\bibitem[Li et~al\mbox{.}(2016)]%
        {li2016study}
\bibfield{author}{\bibinfo{person}{Meicong Li}, \bibinfo{person}{Wei Huang},
  \bibinfo{person}{Yongbin Wang}, \bibinfo{person}{Wenqing Fan}, {and}
  \bibinfo{person}{Jianfang Li}.} \bibinfo{year}{2016}\natexlab{}.
\newblock \showarticletitle{The study of APT attack stage model}. In
  \bibinfo{booktitle}{\emph{2016 IEEE/ACIS 15th International Conference on
  Computer and Information Science (ICIS)}}. IEEE, \bibinfo{pages}{1--5}.
\newblock


\bibitem[Marom and Rosman(2018)]%
        {marom2018belief}
\bibfield{author}{\bibinfo{person}{Ofir Marom} {and} \bibinfo{person}{Benjamin
  Rosman}.} \bibinfo{year}{2018}\natexlab{}.
\newblock \showarticletitle{Belief reward shaping in reinforcement learning}.
  In \bibinfo{booktitle}{\emph{Proceedings of the AAAI conference on artificial
  intelligence}}, Vol.~\bibinfo{volume}{32}.
\newblock


\bibitem[Mazzaglia et~al\mbox{.}(2022)]%
        {mazzaglia2022curiositydriven}
\bibfield{author}{\bibinfo{person}{Pietro Mazzaglia}, \bibinfo{person}{Ozan
  Catal}, \bibinfo{person}{Tim Verbelen}, {and} \bibinfo{person}{Bart Dhoedt}.}
  \bibinfo{year}{2022}\natexlab{}.
\newblock \bibinfo{title}{Curiosity-Driven Exploration via Latent Bayesian
  Surprise}.
\newblock
\newblock
\showeprint[arxiv]{2104.07495}~[cs.LG]


\bibitem[Mirsky and Lee(2021)]%
        {mirsky2021creation}
\bibfield{author}{\bibinfo{person}{Yisroel Mirsky} {and} \bibinfo{person}{Wenke
  Lee}.} \bibinfo{year}{2021}\natexlab{}.
\newblock \showarticletitle{The creation and detection of deepfakes: A survey}.
\newblock \bibinfo{journal}{\emph{ACM Computing Surveys (CSUR)}}
  \bibinfo{volume}{54}, \bibinfo{number}{1} (\bibinfo{year}{2021}),
  \bibinfo{pages}{1--41}.
\newblock


\bibitem[Mnih et~al\mbox{.}(2016)]%
        {mnih2016asynchronous}
\bibfield{author}{\bibinfo{person}{Volodymyr Mnih},
  \bibinfo{person}{Adrià~Puigdomènech Badia}, \bibinfo{person}{Mehdi Mirza},
  \bibinfo{person}{Alex Graves}, \bibinfo{person}{Timothy~P. Lillicrap},
  \bibinfo{person}{Tim Harley}, \bibinfo{person}{David Silver}, {and}
  \bibinfo{person}{Koray Kavukcuoglu}.} \bibinfo{year}{2016}\natexlab{}.
\newblock \bibinfo{title}{Asynchronous Methods for Deep Reinforcement
  Learning}.
\newblock
\newblock
\showeprint[arxiv]{1602.01783}~[cs.LG]


\bibitem[Mnih et~al\mbox{.}(2013a)]%
        {mnih2013playing}
\bibfield{author}{\bibinfo{person}{Volodymyr Mnih}, \bibinfo{person}{Koray
  Kavukcuoglu}, \bibinfo{person}{David Silver}, \bibinfo{person}{Alex Graves},
  \bibinfo{person}{Ioannis Antonoglou}, \bibinfo{person}{Daan Wierstra}, {and}
  \bibinfo{person}{Martin Riedmiller}.} \bibinfo{year}{2013}\natexlab{a}.
\newblock \bibinfo{title}{Playing Atari with Deep Reinforcement Learning}.
\newblock
\newblock
\showeprint[arxiv]{1312.5602}~[cs.LG]


\bibitem[Mnih et~al\mbox{.}(2013b)]%
        {DQN}
\bibfield{author}{\bibinfo{person}{Volodymyr Mnih}, \bibinfo{person}{Koray
  Kavukcuoglu}, \bibinfo{person}{David Silver}, \bibinfo{person}{Alex Graves},
  \bibinfo{person}{Ioannis Antonoglou}, \bibinfo{person}{Daan Wierstra}, {and}
  \bibinfo{person}{Martin~A. Riedmiller}.} \bibinfo{year}{2013}\natexlab{b}.
\newblock \showarticletitle{Playing Atari with Deep Reinforcement Learning}.
\newblock \bibinfo{journal}{\emph{CoRR}}  \bibinfo{volume}{abs/1312.5602}
  (\bibinfo{year}{2013}).
\newblock
\showeprint[arXiv]{1312.5602}
\urldef\tempurl%
\url{http://arxiv.org/abs/1312.5602}
\showURL{%
\tempurl}


\bibitem[Mnih et~al\mbox{.}(2015)]%
        {Mnih2015}
\bibfield{author}{\bibinfo{person}{V. Mnih}, \bibinfo{person}{K. Kavukcuoglu},
  \bibinfo{person}{D. Silver}, \bibinfo{person}{A~A. Rusu}, \bibinfo{person}{J.
  Veness}, \bibinfo{person}{M~G. Bellemare}, \bibinfo{person}{A. Graves},
  \bibinfo{person}{M. Riedmiller}, \bibinfo{person}{A~K. Fidjeland},
  \bibinfo{person}{G. Ostrovski}, \bibinfo{person}{S. Petersen},
  \bibinfo{person}{C. Beattie}, \bibinfo{person}{A. Sadik}, \bibinfo{person}{I.
  Antonoglou}, \bibinfo{person}{H. King}, \bibinfo{person}{D. Kumaran},
  \bibinfo{person}{D. Wierstra}, \bibinfo{person}{S. Legg}, {and}
  \bibinfo{person}{D. Hassabis}.} \bibinfo{year}{2015}\natexlab{}.
\newblock \showarticletitle{{Human-level control through deep reinforcement
  learning}}.
\newblock \bibinfo{journal}{\emph{Nature}} (\bibinfo{year}{2015}).
\newblock


\bibitem[Molina-Markham et~al\mbox{.}(2021)]%
        {molina}
\bibfield{author}{\bibinfo{person}{Andres Molina-Markham},
  \bibinfo{person}{Cory Miniter}, \bibinfo{person}{Becky Powell}, {and}
  \bibinfo{person}{Ahmad Ridley}.} \bibinfo{year}{2021}\natexlab{}.
\newblock \bibinfo{title}{Network Environment Design for Autonomous
  Cyberdefense}.
\newblock
\newblock
\showeprint[arxiv]{2103.07583}~[cs.CR]


\bibitem[Ng et~al\mbox{.}(1999)]%
        {ng1999policy}
\bibfield{author}{\bibinfo{person}{Andrew~Y Ng}, \bibinfo{person}{Daishi
  Harada}, {and} \bibinfo{person}{Stuart Russell}.}
  \bibinfo{year}{1999}\natexlab{}.
\newblock \showarticletitle{Policy invariance under reward transformations:
  Theory and application to reward shaping}. In
  \bibinfo{booktitle}{\emph{Icml}}, Vol.~\bibinfo{volume}{99}. Citeseer,
  \bibinfo{pages}{278--287}.
\newblock


\bibitem[Nyberg and Johnson(2023)]%
        {nyberg2023training}
\bibfield{author}{\bibinfo{person}{Jakob Nyberg} {and} \bibinfo{person}{Pontus
  Johnson}.} \bibinfo{year}{2023}\natexlab{}.
\newblock \bibinfo{title}{Training Automated Defense Strategies Using
  Graph-based Cyber Attack Simulations}.
\newblock
\newblock
\showeprint[arxiv]{2304.11084}~[cs.CR]


\bibitem[OpenAI et~al\mbox{.}(2019)]%
        {openai2019dota}
\bibfield{author}{\bibinfo{person}{OpenAI}, \bibinfo{person}{:},
  \bibinfo{person}{Christopher Berner}, \bibinfo{person}{Greg Brockman},
  \bibinfo{person}{Brooke Chan}, \bibinfo{person}{Vicki Cheung},
  \bibinfo{person}{Przemysław Dębiak}, \bibinfo{person}{Christy Dennison},
  \bibinfo{person}{David Farhi}, \bibinfo{person}{Quirin Fischer},
  \bibinfo{person}{Shariq Hashme}, \bibinfo{person}{Chris Hesse},
  \bibinfo{person}{Rafal Józefowicz}, \bibinfo{person}{Scott Gray},
  \bibinfo{person}{Catherine Olsson}, \bibinfo{person}{Jakub Pachocki},
  \bibinfo{person}{Michael Petrov}, \bibinfo{person}{Henrique~P. d. O.~Pinto},
  \bibinfo{person}{Jonathan Raiman}, \bibinfo{person}{Tim Salimans},
  \bibinfo{person}{Jeremy Schlatter}, \bibinfo{person}{Jonas Schneider},
  \bibinfo{person}{Szymon Sidor}, \bibinfo{person}{Ilya Sutskever},
  \bibinfo{person}{Jie Tang}, \bibinfo{person}{Filip Wolski}, {and}
  \bibinfo{person}{Susan Zhang}.} \bibinfo{year}{2019}\natexlab{}.
\newblock \bibinfo{title}{Dota 2 with Large Scale Deep Reinforcement Learning}.
\newblock
\newblock
\showeprint[arxiv]{1912.06680}~[cs.LG]


\bibitem[Pathak et~al\mbox{.}(2017)]%
        {pathak2017curiosity}
\bibfield{author}{\bibinfo{person}{Deepak Pathak}, \bibinfo{person}{Pulkit
  Agrawal}, \bibinfo{person}{Alexei~A Efros}, {and} \bibinfo{person}{Trevor
  Darrell}.} \bibinfo{year}{2017}\natexlab{}.
\newblock \showarticletitle{Curiosity-driven exploration by self-supervised
  prediction}. In \bibinfo{booktitle}{\emph{International conference on machine
  learning}}. PMLR, \bibinfo{pages}{2778--2787}.
\newblock


\bibitem[Rush et~al\mbox{.}(2015)]%
        {candles}
\bibfield{author}{\bibinfo{person}{George Rush}, \bibinfo{person}{Daniel~R.
  Tauritz}, {and} \bibinfo{person}{Alexander~D. Kent}.}
  \bibinfo{year}{2015}\natexlab{}.
\newblock \showarticletitle{Coevolutionary Agent-Based Network Defense
  Lightweight Event System (CANDLES)}. In \bibinfo{booktitle}{\emph{Proceedings
  of the Companion Publication of the 2015 Annual Conference on Genetic and
  Evolutionary Computation}} (Madrid, Spain) \emph{(\bibinfo{series}{GECCO
  Companion '15})}. \bibinfo{publisher}{Association for Computing Machinery},
  \bibinfo{address}{New York, NY, USA}, \bibinfo{pages}{859–866}.
\newblock
\showISBNx{9781450334884}
\urldef\tempurl%
\url{https://doi.org/10.1145/2739482.2768429}
\showDOI{\tempurl}


\bibitem[Sallab et~al\mbox{.}(2017)]%
        {sallab_deep_2017}
\bibfield{author}{\bibinfo{person}{Ahmad~El Sallab}, \bibinfo{person}{Mohammed
  Abdou}, \bibinfo{person}{Etienne Perot}, {and} \bibinfo{person}{Senthil
  Yogamani}.} \bibinfo{year}{2017}\natexlab{}.
\newblock \showarticletitle{Deep {Reinforcement} {Learning} framework for
  {Autonomous} {Driving}}.
\newblock \bibinfo{journal}{\emph{Electronic Imaging}} \bibinfo{volume}{29},
  \bibinfo{number}{19} (\bibinfo{date}{Jan.} \bibinfo{year}{2017}),
  \bibinfo{pages}{70--76}.
\newblock
\showISSN{2470-1173}
\urldef\tempurl%
\url{https://doi.org/10.2352/ISSN.2470-1173.2017.19.AVM-023}
\showDOI{\tempurl}
\newblock
\shownote{arXiv:1704.02532 [cs, stat]}.


\bibitem[Schoonover et~al\mbox{.}(2018)]%
        {kevin}
\bibfield{author}{\bibinfo{person}{Kevin Schoonover}, \bibinfo{person}{Eric
  Michalak}, \bibinfo{person}{Sean Harris}, \bibinfo{person}{Adam Gausmann},
  \bibinfo{person}{Hannah Reinbolt}, \bibinfo{person}{Daniel~R. Tauritz},
  \bibinfo{person}{Chris Rawlings}, {and} \bibinfo{person}{Aaron~Scott Pope}.}
  \bibinfo{year}{2018}\natexlab{}.
\newblock \showarticletitle{Galaxy: A Network Emulation Framework for
  Cybersecurity}. In \bibinfo{booktitle}{\emph{11th USENIX Workshop on Cyber
  Security Experimentation and Test (CSET 18)}}. \bibinfo{publisher}{USENIX
  Association}, \bibinfo{address}{Baltimore, MD}.
\newblock
\urldef\tempurl%
\url{https://www.usenix.org/conference/cset18/presentation/schoonover}
\showURL{%
\tempurl}


\bibitem[Schulman et~al\mbox{.}(2015)]%
        {schulman2015}
\bibfield{author}{\bibinfo{person}{John Schulman}, \bibinfo{person}{Sergey
  Levine}, \bibinfo{person}{Pieter Abbeel}, \bibinfo{person}{Michael Jordan},
  {and} \bibinfo{person}{Philipp Moritz}.} \bibinfo{year}{2015}\natexlab{}.
\newblock \showarticletitle{Trust region policy optimization}. In
  \bibinfo{booktitle}{\emph{International conference on machine learning}}.
  PMLR, \bibinfo{pages}{1889--1897}.
\newblock


\bibitem[Schulman et~al\mbox{.}(2017a)]%
        {schulman_proximal_2017}
\bibfield{author}{\bibinfo{person}{J. Schulman}, \bibinfo{person}{F. Wolski},
  \bibinfo{person}{P. Dhariwal}, \bibinfo{person}{A. Radford}, {and}
  \bibinfo{person}{O. Klimov}.} \bibinfo{year}{2017}\natexlab{a}.
\newblock \showarticletitle{Proximal {Policy} {Optimization} {Algorithms}}. In
  \bibinfo{booktitle}{\emph{{arXiv}:1707.06347 [cs]}}.
\newblock


\bibitem[Schulman et~al\mbox{.}(2017b)]%
        {PPO}
\bibfield{author}{\bibinfo{person}{John Schulman}, \bibinfo{person}{Filip
  Wolski}, \bibinfo{person}{Prafulla Dhariwal}, \bibinfo{person}{Alec Radford},
  {and} \bibinfo{person}{Oleg Klimov}.} \bibinfo{year}{2017}\natexlab{b}.
\newblock \showarticletitle{Proximal policy optimization algorithms}.
\newblock \bibinfo{journal}{\emph{arXiv preprint arXiv:1707.06347}}
  (\bibinfo{year}{2017}).
\newblock


\bibitem[Schwartz({[n.\,d.]})]%
        {nasim}
\bibfield{author}{\bibinfo{person}{Jonathon Schwartz}.}
  \bibinfo{year}{[n.\,d.]}\natexlab{}.
\newblock \bibinfo{title}{Network Attack Simulator}.
\newblock
  \bibinfo{howpublished}{\url{https://github.com/Jjschwartz/NetworkAttackSimulator},
  Year = {2020}}.
\newblock


\bibitem[Schwartz and Kurniawati(2019)]%
        {schwartz2019autonomous}
\bibfield{author}{\bibinfo{person}{Jonathon Schwartz} {and}
  \bibinfo{person}{Hanna Kurniawati}.} \bibinfo{year}{2019}\natexlab{}.
\newblock \bibinfo{title}{Autonomous Penetration Testing using Reinforcement
  Learning}.
\newblock
\newblock
\showeprint[arxiv]{1905.05965}~[cs.CR]


\bibitem[Singh et~al\mbox{.}(2009)]%
        {Singh}
\bibfield{author}{\bibinfo{person}{Satinder Singh}, \bibinfo{person}{R. Lewis},
  {and} \bibinfo{person}{A. Barto}.} \bibinfo{year}{2009}\natexlab{}.
\newblock \showarticletitle{Where Do Rewards Come From?}
\newblock \bibinfo{journal}{\emph{Proceedings of the 31st Annual Meeting of the
  Cognitive Science Society}}.
\newblock


\bibitem[Surjit(2020)]%
        {Surjit_2020}
\bibfield{author}{\bibinfo{person}{Dr. Surjit}.}
  \bibinfo{year}{2020}\natexlab{}.
\newblock \bibinfo{title}{Deep reinforcement learning using proximal policy
  optimization}.
\newblock
\newblock
\urldef\tempurl%
\url{https://medium.com/analytics-vidhya/deep-reinforcement-learning-using-proximal-policy-optimization-7555280ef941}
\showURL{%
\tempurl}


\bibitem[Sutton and Barto(2018a)]%
        {sutton}
\bibfield{author}{\bibinfo{person}{Richard~S Sutton} {and}
  \bibinfo{person}{Andrew~G Barto}.} \bibinfo{year}{2018}\natexlab{a}.
\newblock \bibinfo{booktitle}{\emph{Reinforcement learning: An introduction}}.
\newblock \bibinfo{publisher}{MIT press}.
\newblock


\bibitem[Sutton and Barto(2018b)]%
        {sutton2018reinforcement}
\bibfield{author}{\bibinfo{person}{Richard~S Sutton} {and}
  \bibinfo{person}{Andrew~G Barto}.} \bibinfo{year}{2018}\natexlab{b}.
\newblock \bibinfo{booktitle}{\emph{Reinforcement learning: An introduction}}.
\newblock \bibinfo{publisher}{MIT press}.
\newblock


\bibitem[Tang et~al\mbox{.}(2017)]%
        {tang2017exploration}
\bibfield{author}{\bibinfo{person}{Haoran Tang}, \bibinfo{person}{Rein
  Houthooft}, \bibinfo{person}{Davis Foote}, \bibinfo{person}{Adam Stooke},
  \bibinfo{person}{OpenAI Xi~Chen}, \bibinfo{person}{Yan Duan},
  \bibinfo{person}{John Schulman}, \bibinfo{person}{Filip DeTurck}, {and}
  \bibinfo{person}{Pieter Abbeel}.} \bibinfo{year}{2017}\natexlab{}.
\newblock \showarticletitle{\# exploration: A study of count-based exploration
  for deep reinforcement learning}.
\newblock \bibinfo{journal}{\emph{Advances in neural information processing
  systems}}  \bibinfo{volume}{30} (\bibinfo{year}{2017}).
\newblock


\bibitem[Team.(2021)]%
        {msft}
\bibfield{author}{\bibinfo{person}{Microsoft Defender~Research Team.}}
  \bibinfo{year}{2021}\natexlab{}.
\newblock \bibinfo{title}{CyberBattleSim}.
\newblock
  \bibinfo{howpublished}{\url{https://github.com/microsoft/cyberbattlesim}}.
\newblock
\newblock
\shownote{Created by Christian Seifert, Michael Betser, William Blum, James
  Bono, Kate Farris, Emily Goren, Justin Grana, Kristian Holsheimer, Brandon
  Marken, Joshua Neil, Nicole Nichols, Jugal Parikh, Haoran Wei.}.


\bibitem[Trott et~al\mbox{.}(2019)]%
        {trott2019keeping}
\bibfield{author}{\bibinfo{person}{Alexander Trott}, \bibinfo{person}{Stephan
  Zheng}, \bibinfo{person}{Caiming Xiong}, {and} \bibinfo{person}{Richard
  Socher}.} \bibinfo{year}{2019}\natexlab{}.
\newblock \showarticletitle{Keeping your distance: Solving sparse reward tasks
  using self-balancing shaped rewards}.
\newblock \bibinfo{journal}{\emph{Advances in Neural Information Processing
  Systems}}  \bibinfo{volume}{32} (\bibinfo{year}{2019}).
\newblock


\bibitem[Wiewiora(2003)]%
        {Wiewiora_2003}
\bibfield{author}{\bibinfo{person}{E. Wiewiora}.}
  \bibinfo{year}{2003}\natexlab{}.
\newblock \showarticletitle{Potential-Based Shaping and Q-Value Initialization
  are Equivalent}.
\newblock \bibinfo{journal}{\emph{Journal of Artificial Intelligence Research}}
   \bibinfo{volume}{19} (\bibinfo{date}{sep} \bibinfo{year}{2003}),
  \bibinfo{pages}{205--208}.
\newblock
\urldef\tempurl%
\url{https://doi.org/10.1613/jair.1190}
\showDOI{\tempurl}


\bibitem[Wolk et~al\mbox{.}(2022)]%
        {wolk2022cage}
\bibfield{author}{\bibinfo{person}{Melody Wolk}, \bibinfo{person}{Andy
  Applebaum}, \bibinfo{person}{Camron Dennler}, \bibinfo{person}{Patrick
  Dwyer}, \bibinfo{person}{Marina Moskowitz}, \bibinfo{person}{Harold Nguyen},
  \bibinfo{person}{Nicole Nichols}, \bibinfo{person}{Nicole Park},
  \bibinfo{person}{Paul Rachwalski}, \bibinfo{person}{Frank Rau}, {and}
  \bibinfo{person}{Adrian Webster}.} \bibinfo{year}{2022}\natexlab{}.
\newblock \bibinfo{title}{Beyond CAGE: Investigating Generalization of Learned
  Autonomous Network Defense Policies}.
\newblock
\newblock
\showeprint[arxiv]{2211.15557}~[cs.LG]


\bibitem[Yang and Liu(2022)]%
        {yang2022behaviourdiverse}
\bibfield{author}{\bibinfo{person}{Yizhou Yang} {and} \bibinfo{person}{Xin
  Liu}.} \bibinfo{year}{2022}\natexlab{}.
\newblock \bibinfo{title}{Behaviour-Diverse Automatic Penetration Testing: A
  Curiosity-Driven Multi-Objective Deep Reinforcement Learning Approach}.
\newblock
\newblock
\showeprint[arxiv]{2202.10630}~[cs.LG]


\bibitem[Yu et~al\mbox{.}(2022)]%
        {yu2022surprising}
\bibfield{author}{\bibinfo{person}{Chao Yu}, \bibinfo{person}{Akash Velu},
  \bibinfo{person}{Eugene Vinitsky}, \bibinfo{person}{Jiaxuan Gao},
  \bibinfo{person}{Yu Wang}, \bibinfo{person}{Alexandre Bayen}, {and}
  \bibinfo{person}{Yi Wu}.} \bibinfo{year}{2022}\natexlab{}.
\newblock \showarticletitle{The surprising effectiveness of ppo in cooperative
  multi-agent games}.
\newblock \bibinfo{journal}{\emph{Advances in Neural Information Processing
  Systems}}  \bibinfo{volume}{35} (\bibinfo{year}{2022}),
  \bibinfo{pages}{24611--24624}.
\newblock


\end{thebibliography}


\end{document}